\documentclass[10pt,twocolumn,letterpaper]{article}
\usepackage[pagenumbers]{cvpr}

\usepackage{caption}
\usepackage{graphicx}
\usepackage{kotex}
\usepackage[T1]{fontenc}      
\usepackage{graphicx}
\usepackage{multirow}
\usepackage{amsmath}          
\usepackage[table]{xcolor}
\usepackage{tabularx}
\usepackage[accsupp]{axessibility}

\newcommand{\myparagraph}[1]{\vspace{1mm}\noindent{\bf #1}}

\definecolor{cvprblue}{rgb}{0.21,0.49,0.74}
\usepackage[pagebackref,breaklinks,colorlinks,allcolors=cvprblue]{hyperref}

\title{FEAST: Fully Connected Expressive Attention for Spatial Transcriptomics}

\author{Taejin Jeong$^{1*}$ \quad Joohyeok Kim$^{1*}$ \quad Jinyeong Kim$^{1}$ \quad Chanyoung Kim$^{1,2}$ \quad Seong Jae Hwang$^{1}$ \\
$^{1}$Yonsei University \quad $^{2}$Emory University \\
{\tt\small \{starforest,kyyle2114,jinyeong1324,seongjae\}@yonsei.ac.kr} \\
{\tt\small chanyoung.kim@emory.edu}
}

\begin{document}
\twocolumn[{%
    \renewcommand\twocolumn[1][]{#1}%
    \maketitle
}]

{\let\thefootnote\relax\footnotetext{* Equal contribution}}

\addtocontents{toc}{\protect\setcounter{tocdepth}{0}}
\begin{abstract}
\noindent{Spatial Transcriptomics (ST) provides spatially-resolved gene expression, offering crucial insights into tissue architecture and complex diseases.} 
However, its prohibitive cost limits widespread adoption, leading to significant attention on inferring spatial gene expression from readily available whole slide images.
While graph neural networks have been proposed to model interactions between tissue regions, their reliance on pre-defined sparse graphs prevents them from considering potentially interacting spot pairs, resulting in a structural limitation in capturing complex biological relationships.
To address this, we propose $\textbf{FEAST}$ (\textbf{F}ully connected \textbf{E}xpressive \textbf{A}ttention for \textbf{S}patial \textbf{T}ranscriptomics), an attention-based framework that models the tissue as a fully connected graph, enabling the consideration of all pairwise interactions. 
To better reflect biological interactions, we introduce {negative-aware attention}, which models both excitatory and inhibitory interactions, capturing essential negative relationships that standard attention often overlooks. 
Furthermore, to mitigate the information loss from truncated or ignored context in standard spot image extraction, we introduce an {off-grid sampling strategy} that gathers additional images from intermediate regions, allowing the model to capture a richer morphological context.
Experiments on public ST datasets show that FEAST surpasses state-of-the-art methods in gene expression prediction while providing biologically plausible attention maps that clarify positive and negative interactions. Our code is available at \url{https://github.com/starforTJ/FEAST}.
\end{abstract}    
\begin{figure*}[t!]
    \centering
    \includegraphics[width=0.95\textwidth]{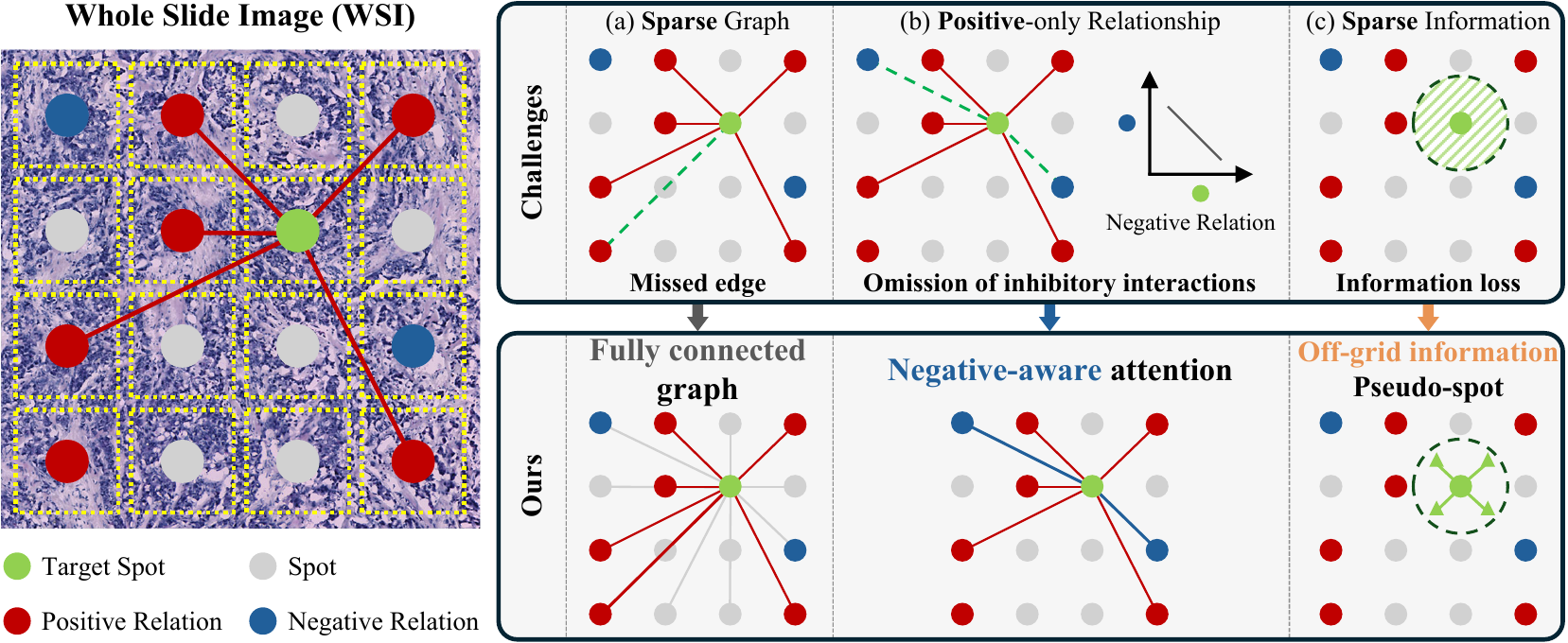}
    \captionof{figure}{Conceptual overview of our proposed framework, illustrating how it addresses key challenges of prior ST methods. Existing approaches are limited by sparse graphs that miss potential interactions, standard positive-only relationships in attention that omit inhibitory interactions, and sparse information that leads to information loss in off-grid regions. Our framework overcomes these limitations by employing a fully connected graph to consider all-pairs interactions, introducing a novel negative-aware attention to explicitly model both positive and inhibitory relationships, and utilizing pseudo-spots to capture and leverage off-grid information.}
    
    \label{fig:teaser}
\end{figure*}

\section{Introduction}
\label{sec:intro}

Spatial Transcriptomics (ST) is a technology that quantifies mRNA expression within its spatial context, thereby enabling the analysis of gene expression while preserving spatial organization~\cite{visium, rao2021exploring, zhang2022clinical}.
Traditional genomic approaches struggle to identify which cells within a tissue interact with one another.
Although single-cell RNA sequencing (scRNA-seq) allows the analysis of RNA expression at the cellular level~\cite{tang2009mrna, hwang2018single}, determining whether such cell-to-cell communication truly occurs within the tissue’s spatial architecture remains challenging.
ST overcomes this limitation by maintaining spatial information during transcriptome profiling, facilitating deeper insights into biologically crucial problems such as tumor microenvironment characterization~\cite{moncada2020integrating, barkley2021recurrence, zuo2024dissecting} and developmental tissue analysis~\cite{srivatsan2021embryo, zhou2023spatial}.
Despite its advantages, acquiring ST data remains highly expensive, making it difficult for individual researchers to obtain routinely.

To address this issue, recent studies have proposed deep learning–based approaches to infer ST data directly from histopathological images~\cite{stnet, histogene, triplex, merge}.
These methods predict spatial mRNA expression profiles from readily available hematoxylin and eosin (H\&E)-stained whole slide images (WSIs).
The fundamental data unit in this task is the \textit{spot},\footnote{To prevent ambiguity, we define our terms as follows: \textit{spot} refers to the ST data unit (i.e., location + expression); \textit{spot image} (or \textit{patch}) refers to the model's image input; and \textit{gene expression} refers to the model's prediction target.} where the goal is to predict its corresponding \textit{gene expression} using the \textit{spot image} (or \textit{patch}).

The underlying premise of these approaches is that \textit{the morphological structure of tissue is closely correlated with its gene expression profile.}
Effectively capturing such structures requires considering not only individual spots but also their relationships, as biological tissue microenvironments arise from complex inter-spot interactions~\cite{armingol2021deciphering, deshmukh2021identification}.
Consequently, mainstream approaches employ graph neural networks (GNNs) built on manually pre-defined sparse graphs. In these graphs, each node represents a spot, and edges connect spots based on pre-defined criteria such as spatial adjacency or morphological feature similarity.

However, pre-defined sparse graphs inherently limit a model’s ability to capture meaningful interactions between spots that are not assigned as neighbors (\Cref{fig:teaser}(a)).
In practice, it is impossible to know \textit{a priori} which spot pairs potentially interact within a tissue; thus, all pairwise interactions should be considered during modeling.
To this end, we advocate a fully connected graph architecture rather than a predefined sparse one.
We further observe that such a fully connected structure can be naturally realized through the attention mechanism, motivating $\textbf{FEAST}$ (\textbf{F}ully connected \textbf{E}xpressive \textbf{A}ttention for \textbf{S}patial \textbf{T}ranscriptomics), an attention-based framework designed to model tissue-wide interactions.
Nonetheless, the standard attention mechanism may fail to fully capture the biological nature of ST data, and existing methods often lose valuable image information.
To overcome these limitations, we introduce two complementary methods.

First, to model complex relationships among spots more expressively, we propose \textit{negative-aware attention} (\Cref{fig:teaser} (b)).
Conventional attention weights are constrained to be non-negative, thereby capturing only similarity or positive correlations between spots.
However, in biological systems, gene interactions can be both excitatory (positive) and inhibitory (negative).
For example, in the tumor microenvironment, the PD-L1/PD-1 immune checkpoint pathway suppresses immune-related gene expression in surrounding cells~\cite{reason1, reason2}.
To represent such duality, our attention formulation allows both positive and negative values without additional computational cost.
This design not only improves predictive performance but also enhances interpretability of the attention mechanism by distinguishing excitatory and inhibitory relationships within the attention map, yielding biologically more plausible explanations.

Second, to mitigate the loss of morphological information caused by patch-based processing, we propose an \textit{off-grid sampling strategy} (\Cref{fig:teaser} (c)).
We observe that biologically informative regions between adjacent spots are often truncated and underutilized due to fixed patch boundaries.
In some cases, spots are separated by large distances, leaving intermediate image regions entirely unused, which further exacerbates information loss.
Our approach samples pseudo-spots from off-grid locations around each original spot and models interactions between original and pseudo-spots.
This strategy enables the model to exploit previously overlooked, information-rich regions of the tissue image. 
To efficiently manage the resulting increase in input images, we employ a \textit{hierarchical attention mechanism}, thereby enhancing both spatial continuity and biological fidelity without excessive computational cost.

Our contributions can be summarized as follows:
\begin{itemize}
    \item We propose FEAST, an attention-based framework that models the tissue as a fully connected graph to overcome the limitations of pre-defined sparse graphs. FEAST introduces \textit{negative-aware attention} to model both excitatory and inhibitory biological relationships.
    \item We introduce an \textit{off-grid sampling strategy} to mitigate critical morphological context lost by standard spot image extraction, and a \textit{hierarchical attention mechanism} to make this approach computationally efficient.
    \item Comprehensive experiments on public ST datasets demonstrate that our framework outperforms current state-of-the-art models in gene expression prediction, while also offering enhanced, biologically plausible interpretability.
\end{itemize}

\section{Related Work}

\begin{figure*}[t!]
    \centering
    \includegraphics[width=\textwidth]{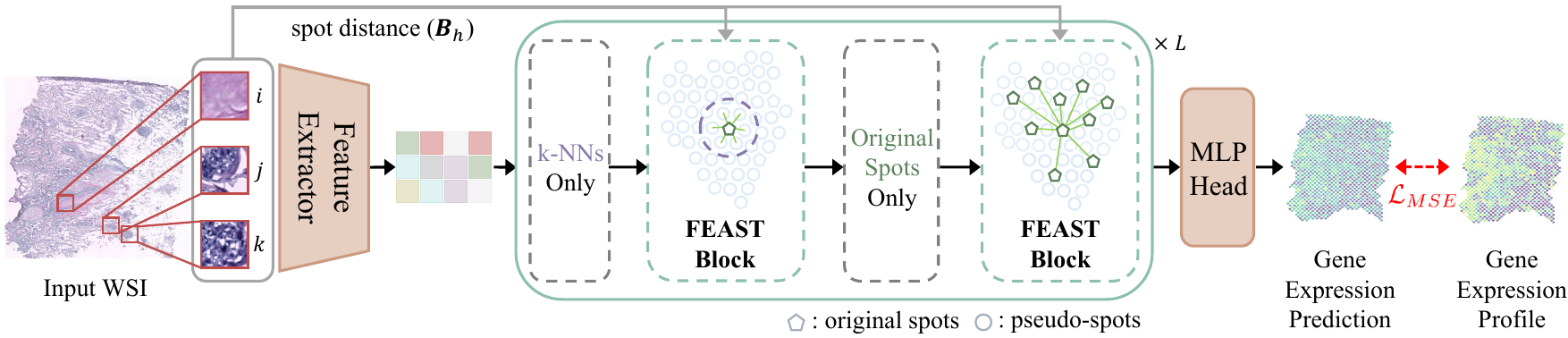}
    \vspace{-15pt}
    \caption{The overall architecture of FEAST. The framework first extracts features and spot distance ($\mathbf{B}_h$) from the input WSI. The features are then processed through $L$ stacked hierarchical attention layers. Each layer consists of two stages: (1) a FEAST Block applied to local $k$-nearest neighbors (including pseudo-spots) to absorb spatial context, followed by (2) a second FEAST Block applied to the original spots for global self-attention. The final representations are passed to an MLP head for gene expression prediction.}
    \label{fig:model architecture}
    \vspace{-10pt}
\end{figure*}

Predicting spatial gene expression profiles from H\&E stained WSIs is a key challenge for understanding the tissue microenvironment. 
This approach is based on the premise that the morphological structure of a tissue is closely correlated with its gene expression profile. 
The initial study, ST-Net~\cite{stnet}, processed each image patch individually using a convolutional neural network. 
This method had a clear limitation as it did not consider interactions between spots, thereby ignoring the spatial context, both local and global.

One direction has been the attempt to capture complex interactions between spots and the overall spatial context using Vision Transformers (ViT)~\cite{vit}, such as in HisToGene~\cite{histogene}. 
However, ST data inherently possesses characteristics of scarcity and heterogeneity. 
This early application of ViT in the ST field was not only vulnerable to these data limitations, but ViT was often applied naively without deeply considering the biologically valid interactions within spatial transcriptomics data. 
These factors combined to limit ViT from fully replicating the high performance seen in the natural image domain within ST data. 
Another approach, TRIPLEX~\cite{triplex}, utilized transformer blocks to extract features from a fixed three-resolution view—the target patch, its neighbors, and the whole slide—and then fused them. 
However, this method had a rigid prior and was limited in its ability to discover flexible, data-driven interactions between distant but morphologically similar spots.

In parallel with these attempts, the dominant research trend has focused on modeling spot interactions using GNNs. 
The initial GNN model, Hist2ST~\cite{hist2ST}, relied on extremely restrictive local connections, such as a 4-nearest-neighbor graph. 
This presented a clear limitation in learning long-range interactions. 
Consequently, subsequent research focused on how to construct a more sophisticated and meaningful graph. 
MERGE~\cite{merge} proposed a hierarchical graph by clustering spots based on two criteria (multi-faceted): (1) spatial proximity and (2) morphological similarity (feature-space), and then creating shortcut edges between them. 
This GNN-based research flow remains the mainstream even today. 
Recently proposed studies~\cite{hyper, thitogene, m2tglgo} continue to advance GNN methodology by introducing hypergraphs, and multi-modal topologies.
All these studies share a common goal of designing more intelligent sparse graphs.
\section{Method}

In this section, we detail the FEAST framework, as illustrated in \Cref{fig:model architecture}.
We begin by revisiting the fundamental limitations of the standard GNN-based methodology for WSI-based gene expression prediction (\cref{sec:limitation of gnn}). 
Next, to overcome these limitations, we introduce an attention-based framework that effectively models the all-pairs interactions between spots (\cref{sec:full graph}). 
We then describe \textit{negative-aware attention}, a novel mechanism designed to explicitly model negative relationships, which are a key aspect of biological interactions that standard attention mechanisms may fail to capture (\cref{sec:neg attn}).
Finally, we propose an \textit{off-grid sampling strategy} to prevent information loss in ST data, and introduce a hierarchical attention mechanism to efficiently manage the resulting increase in computational cost (\cref{sec:off-grid}).

\subsection{Limitations of Sparse Graph Modeling} \label{sec:limitation of gnn}

Given that biological tissue microenvironments arise from complex inter-spot interactions, GNNs have emerged as a natural and widely adopted approach to model these relationships. 
To construct the graphs required for GNNs, these methods rely on the underlying premise that the morphological structure of a tissue is closely correlated with its gene expression profile.
Specifically, edges are typically defined based on two core assumptions: first, \textit{spatial proximity}, which posits that spatially adjacent spots share similar gene expression; and second, \textit{feature similarity}, which assumes that spots with similar morphological features share expression patterns regardless of their spatial distance. 
However, we argue that relying on such pre-defined, sparse graphs presents a fundamental limitation. 
In these structures, connections are restricted to a limited number of neighbors, leaving the vast majority of spot pairs unconnected. 
This implies that any potential biological interaction between these unconnected spots is inevitably ignored by design. 
Consequently, these methods are structurally incapable of fully modeling the complex, tissue-wide interactions where every spot potentially influences every other spot.

\subsection{Attention to Fully Connected Graphs} \label{sec:full graph}

\begin{figure}[t!]
    \centering
    \includegraphics[width=\columnwidth]{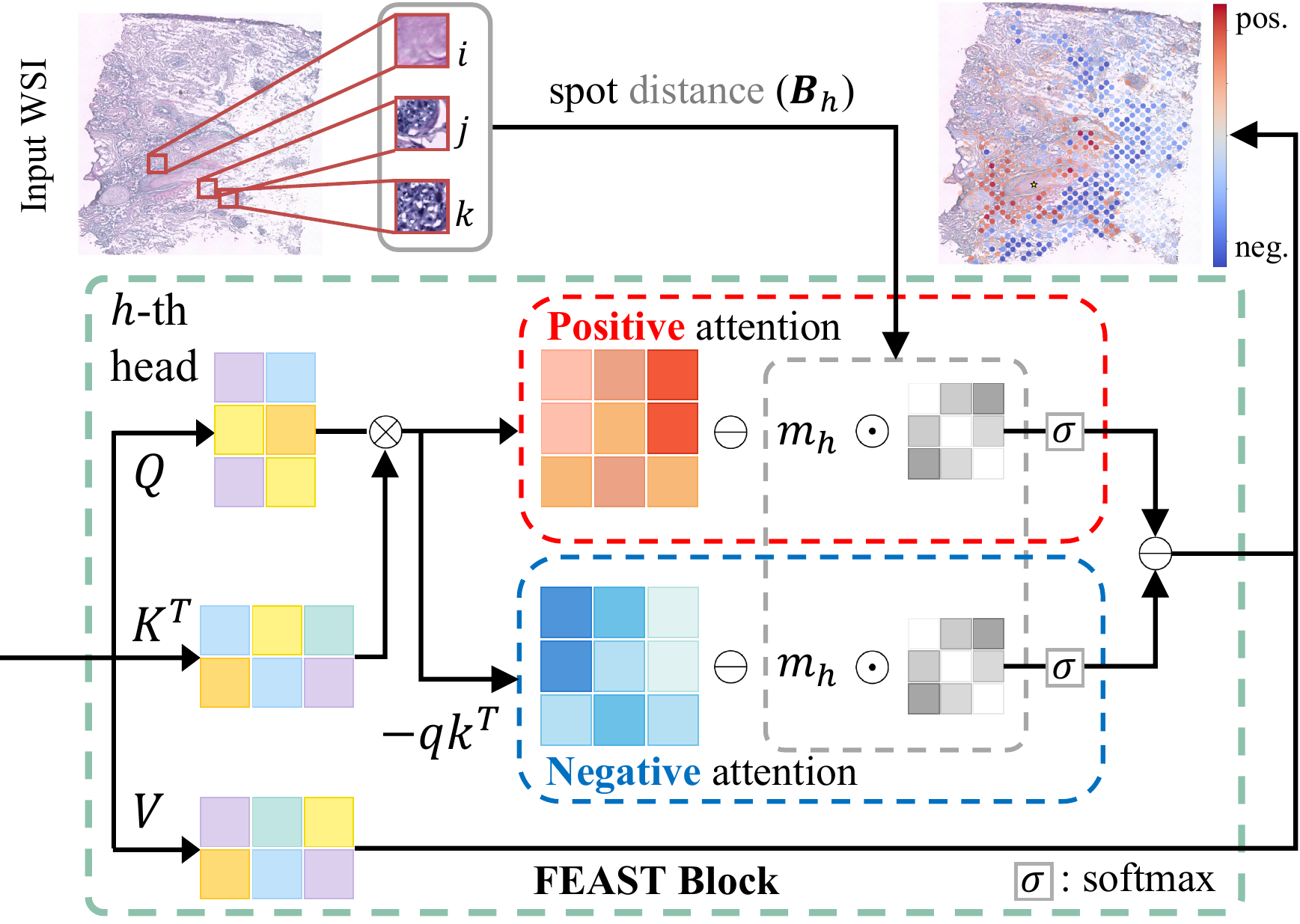}
    \vspace{-15pt}
    \caption{Detailed architecture of the FEAST Block. This block, which serves as the core attention mechanism in our framework, explicitly models both positive (excitatory) and negative (inhibitory) relationships. The raw feature score ($qk^T$) and its inversion ($-qk^T$) are used to compute positive and negative scores, respectively, both incorporating the same positional bias ($\mathbf{B}_h$). After softmax normalization ($\sigma$), the final attention weight is computed by subtracting the negative weight from the positive weight. This results in an attention map that can capture both positive (red) and negative (blue) biological relationships.}
    \label{fig:feast_block}
    \vspace{-10pt}
\end{figure}

To directly address the structural limitation of sparse graphs, we propose FEAST, a framework designed to model the tissue as a fully connected graph and capture all potential pairwise interactions.
Unlike GNNs that rely on pre-defined sparse edges, FEAST leverages the self-attention mechanism to dynamically learn the strength of interactions between every spot pair (\Cref{fig:feast_block}). 
We observe that the inherent architecture of the attention mechanism, particularly the inner product between query ($\mathbf{Q}$) and key ($\mathbf{K}$) matrices, naturally models this fully connected structure.

Given $N$ spots in a WSI, we first obtain the spot feature matrix using a feature extractor. The core of the attention mechanism is to compute the raw attention scores, $\mathbf{S}$, among all $N$ spots via the dot product of $\mathbf{Q}$ and $\mathbf{K}$:
\begin{align}
    \mathbf{S} = \frac{\mathbf{Q}\mathbf{K}^T}{\sqrt{d_k}}, \quad
    \mathbf{A} = \text{softmax}\left(\mathbf{S}\right). \label{eq:attn}
\end{align}
The attention weight matrix, $\mathbf{A}$, can be regarded as a dynamically weighted, fully connected graph that models all relationships between spots. 
Furthermore, the similarity computed by $\mathbf{S}$ serves as a dynamic and learned version of the \textit{feature similarity} criterion, which is a common standard for graph construction for GNNs in this task.

However, the standard attention mechanism in Eq.~\ref{eq:attn} is permutation-invariant, meaning it disregards the spatial arrangement of spots and the distances between them. 
This is suboptimal, as it fails to incorporate the essential assumption that spatially closer spots are more related.
To address this, we incorporate a static, non-learned positional bias, as proposed in ALiBi~\cite{alibi} and TITAN~\cite{titan}.
Concretely, for the $h$-th attention head, a positional bias matrix $\mathbf{B}_h$ is added to the raw attention score $\mathbf{S}_h$ before the softmax operation.
For a spot $i$ at 2D coordinate $(i_x, i_y)$ and a spot $j$ at $(j_x, j_y)$, the bias $\textbf{B}_h(i, j)$ is computed as the Euclidean distance scaled by a head-specific, fixed negative scalar $m_h$:
\begin{align}
  \textbf{B}_h(i, j) = m_h \cdot \sqrt{(i_x - j_x)^2 + (i_y - j_y)^2}.
\end{align}
$m_h$ is a non-learned hyperparameter unique to each head. By assigning a different positional bias, each attention head can model a distinct field of view. Heads with a large $|m_h|$ specialize in local interactions by strongly penalizing attention to distant spots, while heads with an $m_h$ close to 0 can freely learn global, long-range relationships across the tissue. This mechanism allows a single attention block to simultaneously model both local and global interactions. 

\subsection{Negative-aware Attention} \label{sec:neg attn}

Although the attention mechanism allows us to model all-pairs interactions, it still has a limitation in capturing the full complexity of biological relationships.
One of these limitations stems from the standard softmax operation, which constrains the calculated attention weights to a range of $[0.0, 1.0]$.
A high weight indicates a strong positive relationship, while a low weight simply means the spots are irrelevant, signifying an \textit{absence of relation.}

However, this approach is fundamentally insufficient. In complex biological systems, inhibitory (negative) relationships clearly exist, where one factor is actively suppressed by another. The \textit{negative relation} is completely different from the \textit{absence of relation,} yet it has been largely overlooked in prior works. 
The standard softmax-based attention mechanism is incapable of modeling this negative relationship. We, therefore, propose a slight modification to the attention operation to explicitly model both positive and negative relationships.

Our approach is based on the assumption that a spot-to-spot relationship cannot be both positive and negative at the same time. We define the positive attention score ($\mathbf{S}_{\text{pos}, h}$) as the full raw score from the dot product and positional bias. 
To model the negative interactions (i.e., relationships that $\mathbf{S}_{\text{pos}, h}$ would assign low scores to), we define the negative attention score ($\mathbf{S}_{\text{neg}, h}$) as a simple sign inversion of the feature-based dot product:
\begin{align}
    \mathbf{S}_{\text{pos}, h} = \mathbf{S}_h + \textbf{B}_h, \quad
    \mathbf{S}_{\text{neg}, h} = -\mathbf{S}_h + \textbf{B}_h. 
\end{align}
Higher $\mathbf{S}_{\text{pos}, h}$ indicates a strong positive relationship between two spots, whereas spot pairs that had low values in $\mathbf{S}_{\text{pos}, h}$ will now have high values in $\mathbf{S}_{\text{neg}, h}$. This formulation is the simplest way to satisfy our assumption.

These two attention scores are then passed through the softmax function to generate attention weights, $\mathbf{A}_{\text{pos}, h}$ and $\mathbf{A}_{\text{neg}, h}$. 
When calculating the negative attention weight $\mathbf{A}_{\text{neg}, h}$, we introduce a temperature scaling parameter $\tau_{\text{neg}}$:
\begin{align}
    \mathbf{A}_{\text{neg}, h} = \text{softmax}\left(\frac{\mathbf{S}_{\text{neg}, h}}{\tau_{\text{neg}}}\right) .
\end{align}
Here, $\tau_{\text{neg}}$ controls the distribution of the negative influence. We set $\tau_{\text{neg}} < 1$, which sharpens the distribution and forces the model to focus only on the strongest negative relationships.
Lastly, we combine these two weights to calculate the final attention weight, $\mathbf{A}_{\text{final}, h}$, using a hyperparameter $\beta$ to scale the negative contribution:
\begin{align}
    \mathbf{A}_{\text{final}, h} = \mathbf{A}_{\text{pos}, h} - \beta \cdot \mathbf{A}_{\text{neg}, h}.
\end{align}
Unlike standard attention, $\mathbf{A}_{\text{final}, h}$ can now have negative values, enabling our model to capture complex relationships between spots that conventional attention mechanisms may fail to model.

\begin{figure}[t!]
    \centering
    \includegraphics[width=\columnwidth]{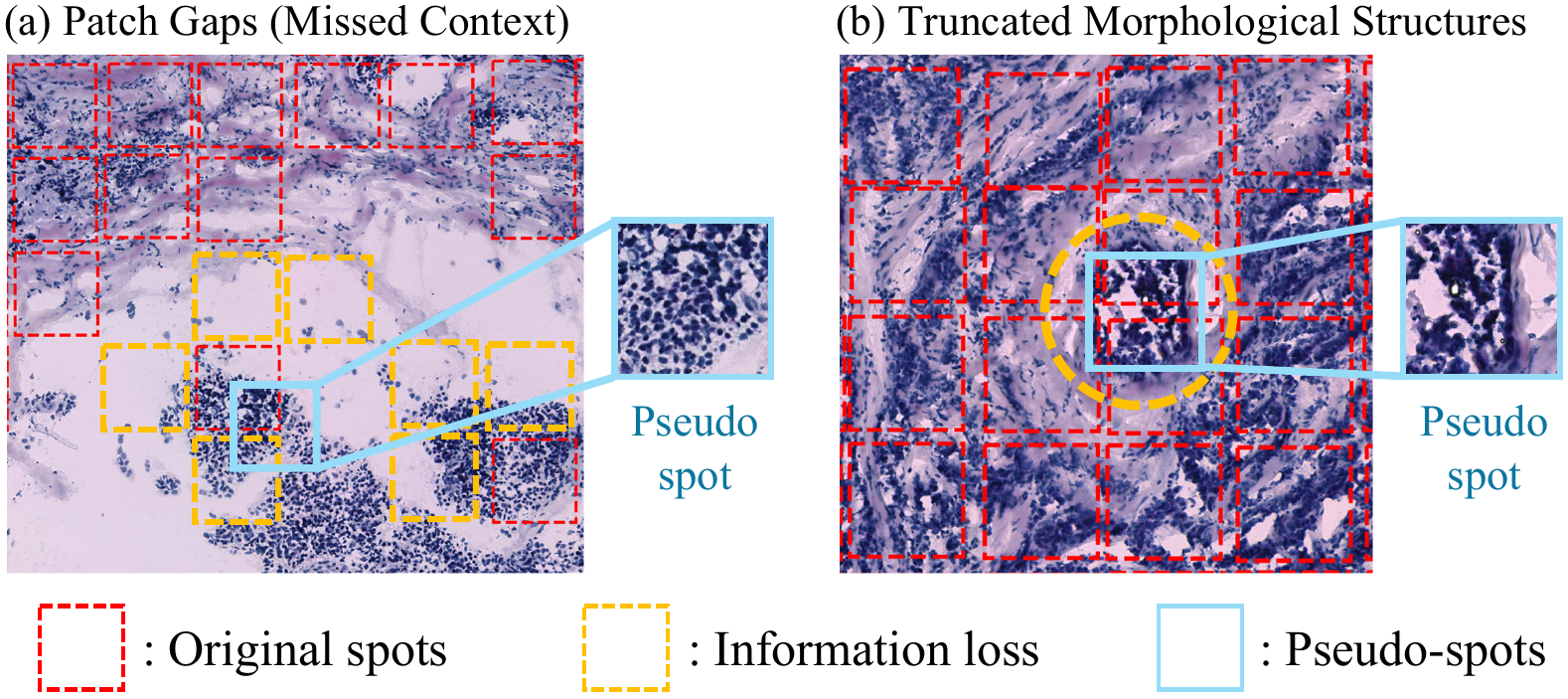}
    \vspace{-15pt}
    \caption{Limitations of the standard fixed-size patch extraction and the motivation for our off-grid sampling. (a) Patch Gaps (Missed Context): Sparse sampling creates off-grid gaps between original patches, causing critical context to be ignored. (b) Truncated Morphological Structures: A single, coherent morphological structure is arbitrarily truncated by the grid boundaries. Our proposed pseudo-spots are sampled from these information-loss regions to restore the complete context.}
    \label{fig:off grid}
    \vspace{-10pt}
\end{figure}

\subsection{Off-grid Sampling Strategy} \label{sec:off-grid}

Most current gene expression prediction frameworks extract fixed-size square patches  centered on the coordinates provided by ST data, which are typically arranged in a grid. However, due to the limitations of ST data acquisition, significant gaps often exist between these spot coordinates. In some cases, the distance between adjacent spots is even larger than the fixed patch size itself. 

Such fixed-size square patch extraction inevitably leads to spatial information loss, as critical morphological structures can be truncated or context from off-grid gaps is ignored, as illustrated in \Cref{fig:off grid}.
While ensuring higher resolution ST data (e.g., Visium HD) might resolve this problem, doing so is often prohibitively expensive. 
Alternatively, one might utilize larger patch sizes to create overlaps between patches. 
However, this strategy introduces a new critical flaw: resizing these large patches to the model's input size causes significant information loss, as morphological details can be distorted or erased in the process. This loss of detail is critical in computational pathology, where such structures are crucial.

To address this problem, we propose an \textit{off-grid sampling strategy}, which introduces pseudo-spots by extracting additional patches from the intermediate locations between the original spot coordinates. 
These pseudo-spots capture critical morphological context that was previously lost. For example, they can complete morphological structures truncated by the original patch boundaries or include image regions that were missed in the gaps. This allows the model to leverage a much richer surrounding context.
However, introducing these pseudo-spots significantly increases the total number of spots ($N$), making the $O(N^2)$ computational cost of the attention mechanism (Eq.~\ref{eq:attn}) computationally unaffordable. To solve this issue, we propose a two-stage \textit{hierarchical attention} block:
\begin{itemize}
    \item Local $k$-NN Attention: In this stage, both original spots and pseudo-spots attend only to their $k$-nearest neighbors. This procedure serves two purposes: it models spot interactions based on spatial proximity, and it allows the original spots to efficiently absorb rich context from their neighboring pseudo-spots.
    \item Global Self-Attention: Self-attention is then applied only to the original spots, which have now been enriched with local context from the previous stage. This step remains computationally affordable, as the attention complexity scales only with the number of original spots, which is significantly smaller than the total number of spots.
\end{itemize}
It is important to note that the attention operation used in both stages is our proposed negative-aware attention (from \cref{sec:neg attn}) that incorporates the positional bias (from Eq.~\ref{eq:attn}). The only difference is the set of spots to which the attention is applied.
After stacking $L$ blocks of this hierarchical attention block, the final representations of the original spots are fed into an MLP head to predict their target gene expression levels. The entire framework is trained using the Mean Squared Error loss ($\mathcal{L}_{\text{MSE}}$).
\renewcommand{\arraystretch}{0.9}
\begin{table*}[t]
\centering
\caption{Cross-validation performance on the ST-Net~\cite{stnet}, Her2ST~\cite{her2st} and SCC~\cite{scc} datasets including baselines. The best results are in \textbf{bold}, and the second best are \underline{underlined}. The performance for all baseline models are cited directly from the MERGE~\cite{merge}.}
\vspace{-3pt}
\label{tab:main_performance_comparison}

\setlength{\tabcolsep}{7pt}

\resizebox{\textwidth}{!}{%
\begin{tabular}{l ccc ccc ccc}

\toprule
\multirow{2}{*}{\textbf{Method}} & \multicolumn{3}{c}{\textbf{ST-Net}} & \multicolumn{3}{c}{\textbf{Her2ST}} & \multicolumn{3}{c}{\textbf{SCC}} \\
\cmidrule(lr){2-4} \cmidrule(lr){5-7} \cmidrule(lr){8-10}

& \rule{1pt}{0ex} MSE $\downarrow$ & MAE $\downarrow$ & PCC $\uparrow$ \rule{1pt}{0ex} & \rule{1pt}{0ex} MSE $\downarrow$ & MAE $\downarrow$ & PCC $\uparrow$ \rule{1pt}{0ex} & \rule{1pt}{0ex} MSE $\downarrow$ & MAE $\downarrow$ & PCC $\uparrow$ \rule{1pt}{0ex} \\

\midrule        
ResNet+FCN & 0.1999 & 0.3448 & 0.5221 & 0.6623 & 0.6385 & 0.4629 & 0.6103 & 0.6290 & 0.4619 \\
BLEEP~\cite{bleep} & 0.3756 & 0.4736 & 0.0784 & 0.7426 & 0.6591 & 0.2747 & 0.6079 & 0.6013 & 0.4176 \\
HisToGene~\cite{histogene} & 0.3054 & 0.4336 & 0.1211 & 0.9452 & 0.7739 & 0.2062 & \textbf{0.3095} & \textbf{0.4367} & 0.1225 \\
Hist2ST~\cite{hist2ST} & 0.3811 & 0.4822 & 0.1525 & 0.7843 & 0.7286 & 0.2479 & 1.0190 & 0.7639 & 0.3003 \\
THItoGene~\cite{thitogene} & 0.2925 & 0.4111 & 0.3666 & 0.8436 & 0.7069 & 0.3445 & 0.6798 & 0.6442 & 0.3897 \\
TRIPLEX~\cite{triplex} & 0.1472 & 0.2943 & 0.2320 & 0.8982 & 0.6946 & 0.3927 & 0.4891 & 0.5356 & 0.5416 \\ 
MERGE~\cite{merge} & \underline{0.1347} & \underline{0.2834} & \underline{0.6795} & \underline{0.6422} & \underline{0.6255} & \underline{0.5037} & 0.5353 & 0.5838 & \underline{0.5512} \\
\midrule 
\rowcolor{blue!3}
FEAST (Ours) & \textbf{0.1177} & \textbf{0.2639} & \textbf{0.7155} & \textbf{0.5761} & \textbf{0.5782} & \textbf{0.5524} & \underline{0.4501} & \underline{0.5239} & \textbf{0.5811} \\
        
\bottomrule
\end{tabular}}
\vspace{-13pt}
\end{table*}

\section{Experiments}
\vspace{-3pt}

We detail the experimental methodology used to evaluate FEAST. All experiments strictly follow the protocol established by MERGE~\cite{merge} to ensure a direct comparison with previous state-of-the-art results.

\myparagraph{Datasets and Preprocessing.} We utilize three public datasets: two breast cancer datasets (ST-Net~\cite{stnet} and Her2ST~\cite{her2st}) and one skin cancer dataset (SCC~\cite{scc}). 
These datasets consist of 68 samples from 23 patients, 36 samples from 8 patients, and 12 samples from 4 patients, respectively. The average number of in-tissue spots per sample is approximately 450, 378, and 723. 
For image preprocessing, spot images are extracted as $256 \times 256$ pixel patches from $\times$20 magnification histopathology images, centered at the ST spot coordinates. 
For gene expression preprocessing, we first follow the protocol by selecting the top 250 most expressed genes for each dataset. The target gene expression profiles are then smoothed using SPCS~\cite{spcs}, the smoothing technique that considers both spatial location and gene expression patterns. 
All results reported hereafter are trained and evaluated on this SPCS-smoothed data. For a more comprehensive description of the preprocessing protocol, we refer the reader to the original MERGE paper.

\myparagraph{Evaluation metrics.} We evaluate generalization performance using an 8-fold cross-validation across all slides. All presented results report the average of these folds. Following prior works, model performance is measured using three metrics: Mean Squared Error (MSE), Mean Absolute Error (MAE), and Pearson Correlation Coefficient (PCC).

\myparagraph{Baselines.} We compare the performance of FEAST against the seven baseline models reported in MERGE~\cite{merge}: ResNet18 with fully connected layer (FCN), BLEEP~\cite{bleep}, HisToGene~\cite{histogene}, Hist2ST~\cite{hist2ST}, THItoGene~\cite{thitogene}, TRIPLEX~\cite{triplex}, and MERGE~\cite{merge}. To ensure a direct and fair comparison, we trained FEAST using the identical experimental protocol, data splits, and preprocessing (SPCS-smoothed data) as MERGE. 

\myparagraph{Implementation details.} We trained a model on a single NVIDIA RTX A6000. 
For the spot image feature extractor, we utilized UNI2-h~\cite{uni} as a fixed   without additional training.
For FEAST, the number of neighbors for the local $k$-NN attention ($k$) was set to 32, and all results are based on this $k$ unless specified otherwise. The hyperparameters for negative-aware attention were set to $\tau_{\text{neg}}=0.6$ and $\beta=1.5$. Additional training details, such as learning rate and optimizer, are provided in the Supplementary Material.

\begin{figure}[t!]
    \centering
    \includegraphics[width=\columnwidth]{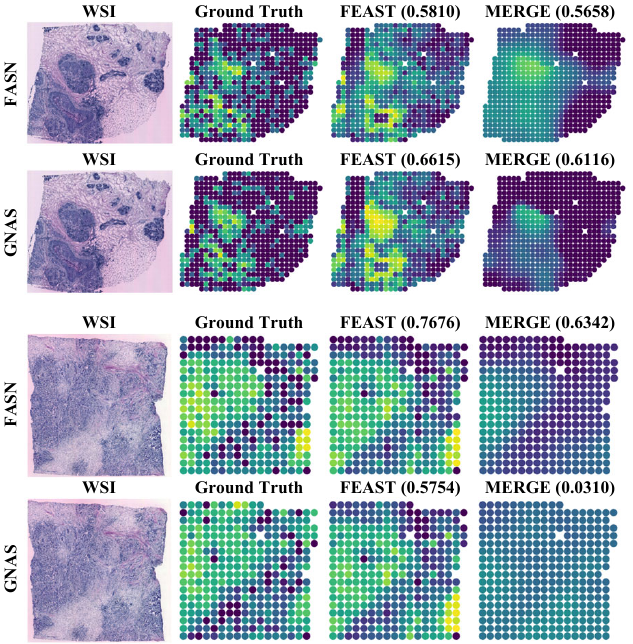}
    \vspace{-15pt}
    \caption{Qualitative comparison of predicted gene expression heatmaps for FEAST and MERGE. Each row presents a side-by-side comparison for the cancer-relevant genes FASN and GNAS, showing the WSI, the ground truth gene expression, and the model predictions. The PCC scores in parentheses confirm that our model attains a higher correlation with the ground truth expressions.}
    \label{fig:qualitative}
    \vspace{-15pt}
\end{figure}

\begin{figure}[t!]
    \centering
    \includegraphics[width=\columnwidth]{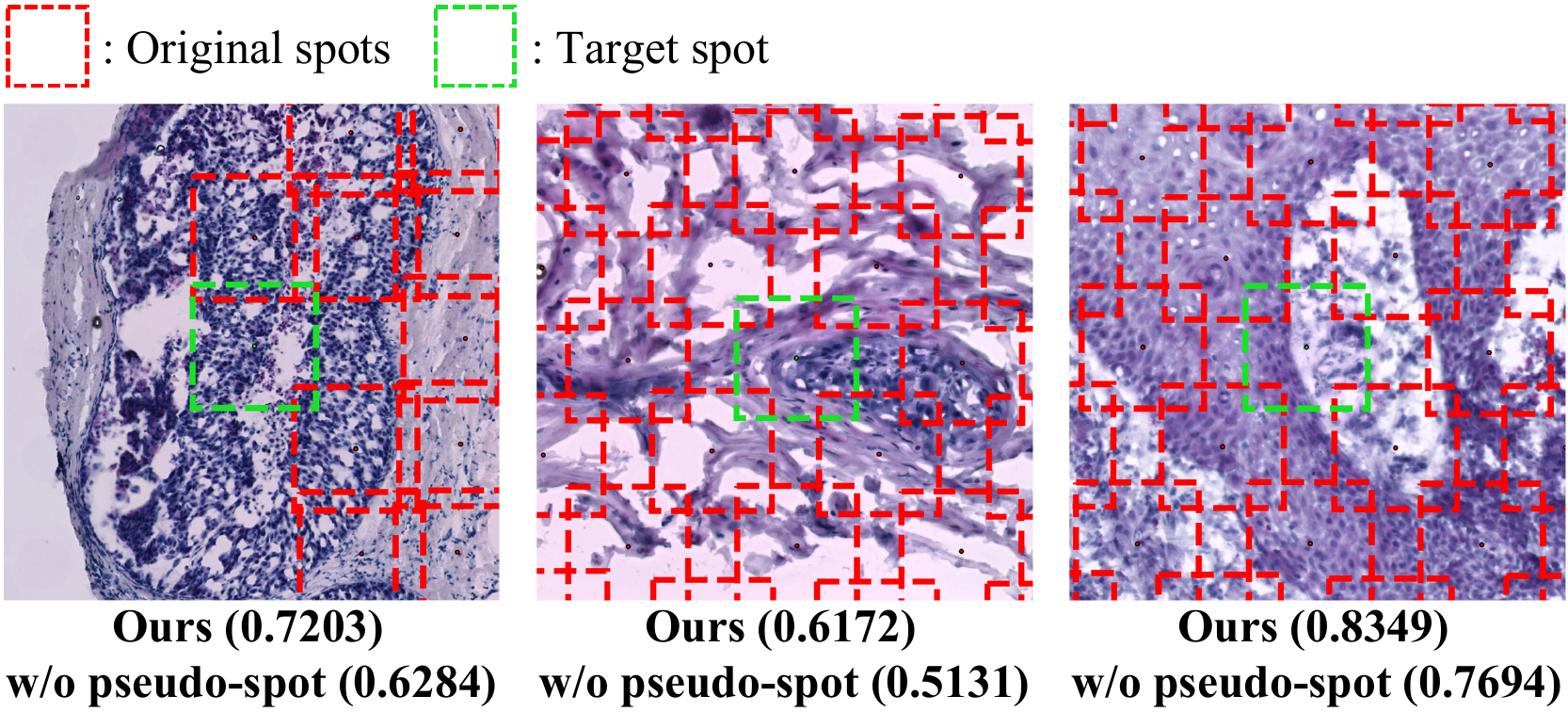}
    \vspace{-15pt}
    \caption{Illustration of the effect of pseudo-spots (off-grid sampling) on target spot prediction. The PCC value in parentheses represents the prediction accuracy for this target spot (green box). The original spots surrounding the target spot (red boxes) are spatially sparse, failing to adequately capture the target's information (w/o pseudo-spot). The PCC value improves when pseudo-spots are incorporated (Ours). This result emphasizes the necessity of our approach when surrounding information is sparse.}
    \label{fig:pseudo_ablation}
    \vspace{-10pt}
\end{figure}

\subsection{Quantitative Results}

\Cref{tab:main_performance_comparison} summarizes the cross-validation performance of FEAST and baseline models across three datasets. 
Our framework achieved overall state-of-the-art performance, recording the best results in 7 out of 9 evaluation metrics and surpassing previous models across all tested datasets.
This dominance was particularly evident in the breast cancer datasets, ST-Net and Her2ST, where our model outperformed all baselines across all metrics. 
The performance gain was most pronounced in the Her2ST dataset, where FEAST achieved an MSE of 0.5761 and a PCC of 0.5524, significantly improving upon the previous best results.
In the SCC dataset, HisToGene recorded the best performance in MSE and MAE. However, FEAST achieved the best performance in PCC with a score of 0.5811. This indicates that our model captures the strongest linear relationship between predicted and actual values, validating its excellent generalization ability in this complex prediction task.

\subsection{Qualitative Results}

Beyond the quantitative metrics, we provide a qualitative analysis to demonstrate the superiority of FEAST. \Cref{fig:qualitative} presents a side-by-side comparison of the predicted gene expression heatmaps from FEAST and MERGE against the ground truth for two cancer-relevant genes, FASN and GNAS~\cite{fasn, gnas}. The heatmaps generated by MERGE appear overly-smoothed and diffuse, failing to capture the fine-grained spatial patterns and sharp boundaries present in the ground truth. 
In contrast, FEAST produces predictions that are visually much closer to the ground truth, accurately restoring the distinct high- and low-expression regions. This visual fidelity is confirmed by the PCC; for instance, in the third row (FASN), FEAST achieves a PCC of 0.7676, whereas MERGE only reaches 0.6342.

To validate the necessity of our off-grid sampling strategy, \Cref{fig:pseudo_ablation} illustrates its direct impact in challenging scenarios characterized by spatial information loss. 
These examples represent cases where the target spot (green box) is surrounded by spatially sparse original spots (red boxes) or where morphological structures are truncated by the fixed patch boundaries. 
In these situations, the model fails to capture the target's full context when pseudo-spots are not used, resulting in significantly lower PCC scores (e.g., 0.6284, 0.5131, 0.7694). 
However, by incorporating pseudo-spots from the surrounding off-grid locations, the model gains access to a much richer morphological context. This leads to a substantial improvement in prediction accuracy (0.7203, 0.6172, and 0.8349, respectively). 
This result confirms that our off-grid sampling strategy is essential for restoring morphological integrity when the surrounding context is sparse or incomplete.

\begin{figure*}[t!]
    \centering
    \includegraphics[width=\textwidth]{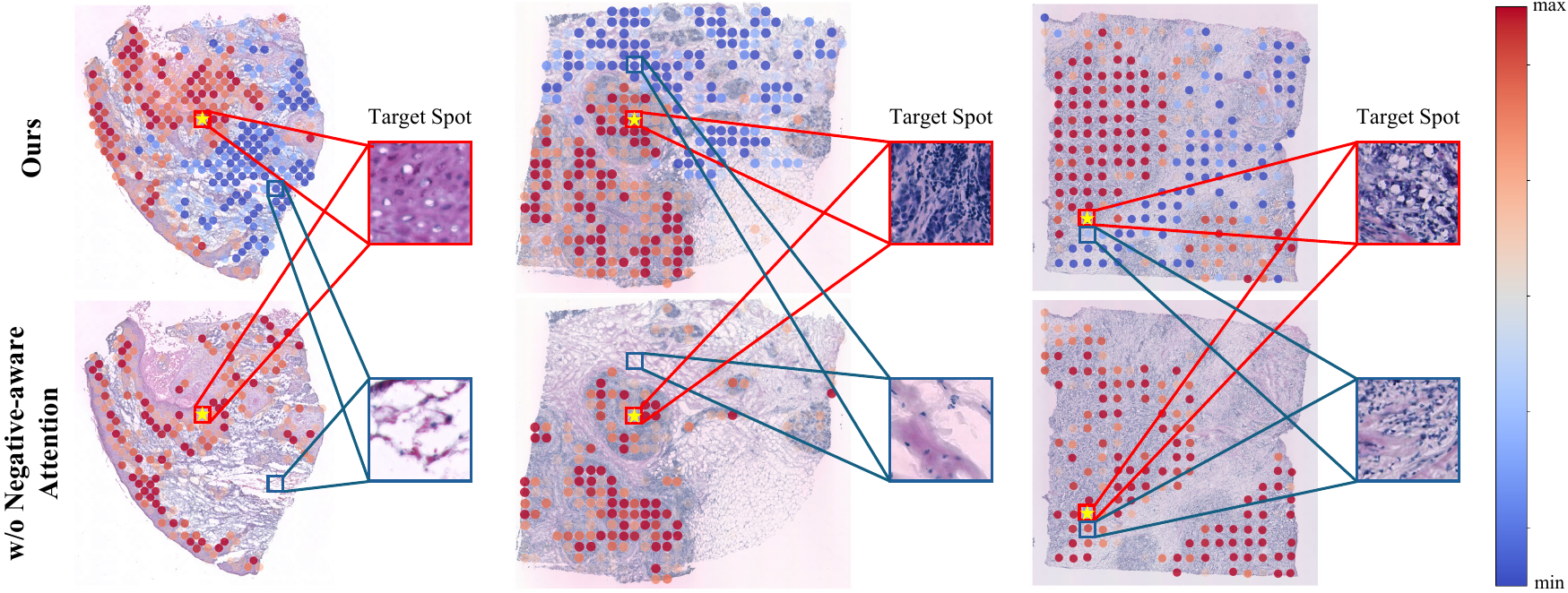}
    \vspace{-15pt}
    \caption{Comparison of attention maps with and without negative-aware attention. Basic attention (bottom) exhibited limitations, either misidentifying structurally different spots as positive or failing to capture them entirely. In contrast, when negative-aware attention was applied (top), it accurately identified structurally dissimilar spots as negative. Notably, we can observe that it clearly distinguishes structural differences even for spatially close spots, where basic attention had failed.}
    \label{fig:attention}
    \vspace{-10pt}
\end{figure*}

\renewcommand{\arraystretch}{0.9}
\begin{table}[t]
\centering
\caption{Performance variation according to the value of $k$. The model achieved the best performance at $k=32$.}
\label{tab:ablation_study_k}
    
\setlength{\tabcolsep}{7pt}

\begin{tabular}{c ccc} 

\toprule
\textbf{$k$} & \rule{1pt}{0ex} MSE $\downarrow$ & MAE $\downarrow$ & PCC $\uparrow$ \rule{1pt}{0ex} \\

\midrule
8 & 0.5821 & 0.5819 & 0.5453 \\
16 & 0.5796 & 0.5784 & 0.5504 \\
32 & \textbf{0.5761} & \textbf{0.5782} & \textbf{0.5524} \\
64 & 0.5793 & 0.5818 & 0.5481 \\
100 & 0.5768 & 0.5807 & 0.5484 \\

\bottomrule
\end{tabular}
\end{table}
\renewcommand{\arraystretch}{0.9}
\begin{table}[t]
\centering
\caption{Ablation study on negative-aware attention and off-grid sampling strategy. The results confirm that applying these methods yields a performance gain compared to when they are not used.}
\label{tab:ablation_study}
    
\setlength{\tabcolsep}{7pt}

\resizebox{\linewidth}{!}{%
\begin{tabular}{cc ccc}

\toprule
{Negative} & {Off-grid} & \rule{1pt}{0ex} MSE $\downarrow$ & MAE $\downarrow$ & PCC $\uparrow$ \rule{1pt}{0ex} \\

\midrule
 & & 0.5878 & 0.5875 & 0.5396 \\
\checkmark & & 0.5778 & 0.5829 & 0.5464 \\
 & \checkmark & 0.5829 & 0.5831 & 0.5458 \\
\checkmark & \checkmark & \textbf{0.5761} & \textbf{0.5782} & \textbf{0.5524} \\
        
\bottomrule
\end{tabular}}
\end{table}

\subsection{Ablation Study}

In this section, we analyze the impact of our model's key design elements. 
We conducted an ablation study to verify: (1) the impact of $k$ in Local $k$-NN Attention, (2) the contribution of pseudo-spots, and (3) the effect of the negative-aware attention mechanism. For brevity, we report the results on the Her2ST dataset.
Full results for all three datasets are available in the Supplementary Material.

\myparagraph{Impact of $k$ in Local $k$-NN Attention.}
The value of $k$ in Local $k$-NN Attention is a key hyperparameter that determines how much surrounding information the target spot references. We analyze its impact using the Her2ST dataset, with results shown in \Cref{tab:ablation_study_k}. 
Performance improves as $k$ increases, reaching its peak at $k=32$ across all evaluation metrics. Beyond this point, further increasing $k$ leads to a decline in performance. Consequently, we selected $k=32$ as the optimal value for FEAST, as it provides the most robust results while ensuring computational efficiency. 

\myparagraph{Effect of Pseudo-spots and Negative Attention.}
Based on the preceding analysis, we set the local window size to $k=32$ and evaluated the contribution of our core components: off-grid sampling (pseudo-spots) and negative-aware attention. 
\Cref{tab:ablation_study} shows the performance change on the Her2ST dataset when each component is ablated. The baseline model (without either component) achieved a PCC of 0.5396. Applying only negative-aware attention or only off-grid sampling offered slight variations, but the full FEAST model, incorporating both components, achieved the best performance across all metrics (PCC 0.5524). These results confirm that both elements are essential contributors and achieve optimal results when used in combination.

\subsection{Attention Map Analysis}

The negative-aware attention introduced in \Cref{sec:neg attn} enhances the representational capacity of the model by capturing both excitatory and inhibitory relationships.
Such relationships are represented through both positive and negative attention weights, reflecting the dual nature of interactions inherent in biological systems.
In \Cref{fig:attention}, we compare the attention weights for the same target spot with and without negative-aware attention.

Without negative-aware attention, the attention heads primarily assign high weights to regions with similar cellularity (\eg, normal or cancer cell regions) and low weights to tissue areas with different characteristics.
In contrast, when negative-aware attention is applied, the model not only assigns positive weights to biologically similar regions but also assigns negative weights to regions with distinct gene expression patterns, such as adipose tissue~\cite{her2st}.
This indicates that the model incorporates inhibitory relationships to refine the representation of the target spot.
Notably, in the last example of \Cref{fig:attention}, a nearby spot that exhibits strong negative correlation with the target spot, potentially providing important complementary information, is almost ignored by standard attention. 

In comparison, our negative-aware attention successfully models this relationship through negative attention weights, highlighting its ability to capture biologically meaningful inhibitory interactions.
Additional examples are provided in the Supplementary Material.
These findings demonstrate that negative-aware attention more faithfully reflects biological relationships and offers richer interpretability by distinguishing excitatory and inhibitory patterns within tissue structures.
\section{Conclusion}

We present FEAST, an attention-based framework that overcomes the limitations of pre-defined sparse graphs in Spatial Transcriptomics by modeling tissues as fully connected graphs and considering all pairwise interactions. 
At the core of our method, FEAST introduces negative-aware attention to explicitly capture both excitatory and inhibitory relationships, and employs an off-grid sampling strategy to recover morphological context lost in standard spot extraction. 
Consequently, FEAST achieves state-of-the-art on gene expression prediction and offers enhanced interpretability via biologically plausible attention maps that distinguish positive and negative interactions.
Furthermore, beyond its favorable performance, FEAST provides a principled basis for quantifying inhibitory effects, paving the way for downstream biological validation and hypothesis generation.
\clearpage
\myparagraph{Acknowledgement.} This work was supported in part by the IITP RS-2024-00457882 (AI Research Hub Project), IITP 2020-II201361, NRF RS-2024-00345806, NRF RS-2023-002620, RQT-25-120390, a Korea Basic Science Institute (National Research Facilities and Equipment Center) grant funded by the Ministry of Science and ICT (No. RS-2024-00403860), and the Advanced Database System Infrastructure (NFEC-2024-11-300458), while the authors are affiliated with the Department of Artificial Intelligence (J.K., C.K., S.J.H.) and the Department of Computer Science (T.J.).
{
    \small
    \bibliographystyle{ieeenat_fullname}
    \bibliography{main}
}

\addtocontents{toc}{\protect\setcounter{tocdepth}{2}}
\appendix
\setcounter{page}{1}
\maketitlesupplementary

\tableofcontents

\vspace{10pt}
In the supplementary material, we begin with implementation details in \Cref{sec:details}. \Cref{sec:details_setup} describes the experimental setup, including hardware specifications and hyperparameters. \Cref{sec:details_preprocess} details the data pre-processing steps, including the smoothing technique and target gene sets.
\Cref{sec:analysis} presents additional experimental analyses. \Cref{sec:analysis_ablation} provides the full ablation study results across all benchmark datasets. \Cref{sec:analysis_elements} covers the analysis of various hyperparameters (e.g., $k$, $\tau_{\text{neg}}$, $\beta$) and design choices, such as feature extractors.
Finally, \Cref{sec:visualization} presents additional qualitative visualizations, including detailed comparisons of PCC distributions and gene expression heatmaps.

\section{Implementation Details} \label{sec:details}

\subsection{Experimental Setup} \label{sec:details_setup}

\myparagraph{Framework and Hardware.} Our framework was implemented using PyTorch and trained on a single NVIDIA RTX A6000 GPU. To ensure reproducibility and a fair comparison, we fixed the random seed to 3927, identical to the seed used in the official MERGE~\cite{merge} implementation.

\myparagraph{Off-grid Sampling Implementation.} The original ST data provides integer grid coordinates (row, col) and their corresponding physical WSI coordinates. Our off-grid sampling strategy generates pseudo-spots at intermediate floating-point grid coordinates (e.g., extracting a spot at $17.5 \times 25$, between $17 \times 25$ and $18 \times 25$). Specifically, the generation process involves:
\begin{enumerate}
    \item Computing the affine transformation matrix to map grid coordinates to physical WSI coordinates.
    \item Generating candidate pseudo-spot coordinates at intermediate locations between the original grid coordinates.
    \item Calculating the average nearest-neighbor distance among original spots.
    \item Filtering candidates: To prevent sampling from background regions, we discard pseudo-spots whose distance to the nearest original spot exceeds the threshold derived in step 3.
\end{enumerate}

\begin{table}[t!]
\centering
\caption{Hyperparameter configuration for FEAST training.}
\label{tab:hyperparams}
\resizebox{\columnwidth}{!}{%
\begin{tabular}{lc}
\toprule
\textbf{Config} & \textbf{Value} \\
\midrule
\multicolumn{2}{c}{\textit{Architecture}} \\
\midrule
Input dimension ($d$) & 1536 \\
Number of attention heads (H) & 8 \\
Number of hierarchical layers ($L$) & 3 \\
Local neighborhood size ($k$) & 32 \\
\midrule
\multicolumn{2}{c}{\textit{Negative-aware Attention}} \\
\midrule
Negative scale factor ($\beta$) & 1.5 \\
Negative temperature ($\tau_{\text{neg}}$) & 0.6 \\
Positional bias slope ($m_h$) & $-2^{-1}, -2^{-2}, \dots, -2^{-H}$ \\
\midrule
\multicolumn{2}{c}{\textit{Optimization}} \\
\midrule
Optimizer & Adam \\
Learning rate & $1 \times 10^{-4}$ \\
Weight decay & $1 \times 10^{-5}$ \\
LR scheduler & Cosine Annealing \\
Epochs & 1500 \\
\bottomrule
\end{tabular}%
}
\end{table}

\begin{table*}[t]
\centering
\caption{List of 250 target genes selected for each dataset.}
\label{tab:gene_list}

\renewcommand{\tabularxcolumn}[1]{m{#1}}

\footnotesize
\renewcommand{\arraystretch}{1.3} 

\begin{tabularx}{\textwidth}{>{\centering\arraybackslash\bfseries}m{1.5cm}X}
\hline

Dataset& 
\multicolumn{1}{>{\centering\arraybackslash}X}{\textbf{250 genes (Total genes to be predicted)}} \\ 
\hline

ST-Net & 
RPS3, IGLL5, RPLP1, TFF3, RPS18, GAPDH, TMSB10, RPLP2, RPS14, RPL37A, RPS19, RPL28, KRT19, RPL8, RPL13, RPL19, ACTB, RPL36, RPL18A, RPL35, RPL18, RPS2, RPS12, RPS21, RACK1, RPL13A, CTSD, FTL, PFN1, MGP, RPS15, RPS11, RPS16, HLA-B, UBA52, NHERF1, RPS17, PSAP, RPLP0, SERF2, RPS27, RPS8, RPL27A, MUC1, RPS28, H2AJ, RPL10, CALR, RPS29, RPL38, RPL11, P4HB, RPS6, CST3, FTH1, RPS4X, SSR4, RPL30, ERBB2, APOE, AZGP1, RPL3, COX6C, HLA-C, FAU, RPS9, EEF2, B2M, RPS5, RPL12, ACTG1, RPS27A, RPL37, RPL23, HLA-A, RPL31, RPL29, RPL7A, IFI27, PABPC1, CD74, BEST1, RPL32, FASN, S100A9, GPX4, RPL15, RPL27, MZT2B, RPL23A, HSPB1, MALAT1, RPS24, COL1A1, C4B, KRT18, CFL1, CD81, ALDOA, RPL35A, SYNGR2, PPP1CA, HLA-E, TAGLN, RPL9, CD63, RPS3A, LGALS3BP, IGFBP2, BST2, TPT1, EDF1, RPS25, ATP6V0B, TAPBP, GRINA, XBP1, S100A11, NBEAL1, AEBP1, CCND1, OAZ1, RPL14, TAGLN2, FN1, PPDPF, BCAP31, IFITM3, PRDX1, BGN, GNAS, PTMA, UBC, MZT2A, SLC25A6, RPS20, HSP90AB1, RPS10, MYL6, CLDN3, ATP6AP1, PRDX2, RPL24, GNB2, RPL34, RPL4, LMNA, NDUFA13, HLA-DRA, SNHG25, TIMP1, H1-0, RPS23, COX8A, KRT8, LY6E, ENO1, GRN, PTPRF, RPL7, UBB, BSG, ELOB, COX6B1, TMSB4X, C1QA, PRSS8, RPL5, UQCR11, RPS7, A2M, RPS15A, VIM, S100A6, NDUFA11, PSMD3, EVL, APOC1, H3-3B, ATP5F1E, PLXNB2, MYL9, TUBA1B, CTSB, ISG15, FLNA, RPS13, NDUFB9, EIF4A1, POLR2L, CYBA, CRIP2, EEF1D, ATP1A1, ELF3, TUFM, SH3BGRL3, STARD10, C3, GUK1, ZNF90, C12orf57, TLE5, SEC61A1, SDC1, PLD3, SPDEF, ARHGDIA, IFI6, LAPTM5, RPL41, CLU, GNAI2, PFDN5, RPL39, SSR2, COX4I1, RHOC, JUP, EIF4G1, FXYD3, TSPO, UQCRQ, COL1A2, RPL10A, S100A8, SELENOW, TPI1, ATP5MC2, PTMS, IGFBP5, LGALS1, SPINT2, RPSA, GSTP1, CHCHD2, EIF5A, COX5B, ATG10, RPL6, EEF1A1, CAPNS1, LMAN2, UBE2M, SPARC, EIF3C, GAS5, TUBB, ACTN4, IGFBP4 \\
\hline

Her2ST & 
IGKC, TMSB10, ERBB2, IGHG3, IGLC2, IGHA1, GAPDH, ACTB, IGLC3, IGHM, SERF2, PSMB3, PFN1, ACTG1, KRT19, RACK1, MUCL1, CISD3, APOE, MIEN1, SSR4, CALR, PSAP, CTSD, FTL, FTH1, TPT1, PTPRF, UBA52, P4HB, BEST1, HLA-B, FAU, SLC9A3R1, FN1, COL1A1, EEF2, IGHG4, CALML5, CD74, B2M, FASN, S100A9, MGP, CFL1, PSMD3, IGHG1, HLA-A, S100A6, MYL6, COL1A2, PHB, TAGLN2, HLA-E, HLA-C, KRT7, CD63, SYNGR2, STARD3, PABPC1, GPX4, GRB7, SLC25A6, AEBP1, GNAS, NDUFB9, EDF1, CRIP2, DDX5, OAZ1, EIF4G1, LMNA, GNB2, CST3, PCGF2, SDC1, S100A11, PRDX1, GRINA, ATP6V0B, TFF3, HLADRA, EEF1D, AZGP1, PPP1CA, FLNA, COL3A1, ATP5E, SPDEF, AP000769.1, ALDOA, PLXNB2, TAGLN, TUBA1B, APOC1, PRRC2A, LAPTM5, PTMS, KRT18, IFI27, PLD3, ADAM15, C1QA, AES, TSPO, MLLT6, TAPBP, SCAND1, ATP1A1, CD81, SEC61A1, CLDN3, PPDPF, S100A14, BGN, C3, MZT2B, S100A8, MDK, PFDN5, H2AFJ, SH3BGRL3, ENO1, XBP1, CYBA, COX6B1, TRAF4, CD24, PRSS8, MMP14, MUC1, VIM, MIDN, SPINT2, BST2, TIMP1, GUK1, ACTN4, CTSB, COX4I1, CCT3, HNRNPA2B1, SEPW1, LY6E, SCD, HSPB1, EIF4G2, BSG, ZYX, TUBB, LASP1, CD99, COL6A2, H1FX, RALY, UBE2M, SPARC, ATG10, HSP90AB1, ORMDL3, LMAN2, CHCHD2, COX7C, ARHGDIA, VMP1, UBC, IGFBP2, COPE, NUPR1, PERP, KRT81, PPP1R1B, LGALS3BP, SSR2, KIAA0100, MYL9, CIB1, IDH2, STARD10, LGALS1, COX6C, GRN, MAPKAPK2, GNAI2, KDELR1, COL18A1, UQCRQ, COX5B, ELOVL1, CHPF, CLDN4, C12orf57, LGALS3, HSP90AA1, JUP, A2M, NDUFB7, PGAP3, HSPA8, TCEB2, PEBP1, COPS9, ATP5G2, ATP6AP1, MYH9, LSM4, COX8A, UQCR11, ATP5B, DHCR24, PTBP1, EIF3B, NDUFA3, FKBP2, MMACHC, RABAC1, ISG15, PTMA, RRBP1, POSTN, C1QB, BCAP31, PSMB4, LAPTM4A, INTS1, FNBP1L, JTB, NBL1, HM13, SLC2A4RG, ROMO1, SERINC2, NDUFA11, RHOC, TXNIP, TYMP, NACA, HSP90B1, SNRPB, PFKL, VCP, ERGIC1, NUCKS1, PSMD8, CALM2, AP2S1, DBI, C4orf48, SDF4, TPI1 \\ 
\hline

SCC & 
S100A8, KRT6A, KRT14, S100A9, KRT5, KRT6B, KRT16, KRT6C, KRT17, MT-CO3, S100A7, MT-CO2, SFN, S100A2, MT-CO1, ACTB, PERP, SPRR1B, KRT10, KRT1, EEF1A1, RPLP1, LGALS7B, LGALS7, COL1A1, FABP5, RPS12, HLA-B, MT-ND4, RPLP2, ACTG1, GJB2, B2M, TPT1, RPL13, MT-ATP6, RPS24, PFN1, KRTDAP, RPS6, DMKN, RPLP0, MT-ND3, RPL37A, DSP, CXCL14, RPS18, RPS17, RPS8, RPL13A, MT-CYB, RPL11, RPL27A, RPL28, MT-ND1, RPS27, RPL32, CSTA, RPL34, RPL31, COL1A2, RPL8, SBSN, TMSB10, ENO1, RPS14, RPL36, SPRR2A, RPL39, GSTP1, RPS27A, JUP, RPS19, RPL37, RPL27, RPL3, RPS29, COL3A1, RPS11, CSTB, RPL9, RACK1, ANXA2, RPL7A, RPL23, RPL19, S100A11, RPS2, RPS28, EEF2, ANXA1, CD74, PABPC1, LDHA, RPS3, RPL35A, DSC2, AQP3, RPS25, IFI27, CALML5, YWHAZ, RPL6, TMSB4X, RPS23, RPL12, S100A14, RPS4X, UBA52, SLPI, PKP1, RPL38, HLA-A, RPS13, LY6D, RPL24, ATP1B3, MYL6, GJB6, S100A6, HSPB1, RPL18, MT-ND2, SDC1, IVL, FTL, RPS3A, RPL10, RPS15A, PI3, RPL18A, S100A10, RPS7, S100A7A, RPL29, RPL26, RPL41, RPL4, RPL7, SPARC, VIM, PTMA, RPS20, MMP1, SH3BGRL3, RPL15, MYH9, GJA1, ITM2B, PPIA, RPL14, UBC, RPL5, CD44, AHNAK, RPL21, DSC3, CNFN, CD24, CFL1, COL17A1, HSP90AA1, RPS16, PKM, NACA, RPS5, ALDOA, H3F3B, S100A16, TAGLN2, HLA-C, TRIM29, LYPD3, FAU, LMNA, SPINK5, SPRR2E, RPL22, KRT2, CST3, DSG3, CLCA2, RPSA, DSG1, RPS9, NDRG1, AC090498.1, GRN, TXN, HSPA8, TGFBI, CTSB, SPRR2D, HLA-DRA, ACTN4, RPS21, EIF1, CTSD, ARPC2, CALML3, KLK7, CALM1, GNAS, DYNLL1, FLG, FLNA, DST, SLC2A1, PSAP, EIF4G2, EEF1B2, FGFBP1, LGALS1, ITGA6, MYL12B, TPI1, RPL10A, TMEM45A, BTF3, DSTN, RTN4, HNRNPA2B1, LAD1, ATP1A1, SERPINB3, PRDX1, COL6A1, ATP5E, PPDPF, TYMP, CD63, EIF5A, YWHAQ, PGK1, HLA-E, IFITM3, RPS26, IGFBP4, OAZ1, NPM1, LCE3D, FXYD3, MT2A, COL6A2, POLR2L, CD59, HNRNPK, RPL35, TMBIM6, HSP90AB1 \\ 
\hline

\end{tabularx}
\end{table*}

\myparagraph{FEAST Training Workflow.} To optimize training efficiency, we pre-extracted image embeddings using UNI2-h~\cite{uni} for both original and pseudo-spots. Let $N$ denote the total number of spots (original + pseudo) and $d$ the feature dimension. For each WSI slide, we saved the features as a numpy array of shape $(N, d)$. During the training of the FEAST framework, we directly load these pre-computed \texttt{.npy} files rather than processing raw images in real-time.

\myparagraph{Hyperparameters.} We provide a detailed summary of the hyperparameter configuration used for training FEAST in \Cref{tab:hyperparams}. 
All experimental results reported in the main text were obtained using this specific configuration.

\renewcommand{\arraystretch}{0.9}
\begin{table*}[t]
\centering
\caption{Ablation study on negative-aware attention and off-grid sampling strategy. The best results are in \textbf{bold}.}
\label{tab:sup_ablation_study}

\setlength{\tabcolsep}{7pt}

\resizebox{\textwidth}{!}{%
\begin{tabular}{cc ccc ccc ccc}

\toprule
\multirow{2}{*}{Negative} & \multirow{2}{*}{Off-grid} & \multicolumn{3}{c}{\textbf{ST-Net}} & \multicolumn{3}{c}{\textbf{Her2ST}} & \multicolumn{3}{c}{\textbf{SCC}} \\
\cmidrule(lr){3-5} \cmidrule(lr){6-8} \cmidrule(lr){9-11}

& & \rule{1pt}{0ex} MSE $\downarrow$ & MAE $\downarrow$ & PCC $\uparrow$ \rule{1pt}{0ex} & \rule{1pt}{0ex} MSE $\downarrow$ & MAE $\downarrow$ & PCC $\uparrow$ \rule{1pt}{0ex} & \rule{1pt}{0ex} MSE $\downarrow$ & MAE $\downarrow$ & PCC $\uparrow$ \rule{1pt}{0ex} \\

\midrule
 & & 0.1156 & 0.2700 & 0.7200 & 0.6538 & 0.6505 & 0.4920 & 0.5010 & 0.5730 & 0.5908 \\

\checkmark & & 0.1144 & 0.2696 & \textbf{0.7267} & 0.6321 & 0.6401 & 0.4958 & 0.4955 & 0.5522 & \textbf{0.5978} \\

 & \checkmark & 0.1126 & 0.2607 & 0.7222 & 0.6308 & 0.6397 & 0.5020 & 0.5035 & 0.5575 & 0.5850 \\

\checkmark & \checkmark & \textbf{0.1111} & \textbf{0.2603} & 0.7224 & \textbf{0.6158} & \textbf{0.6342} & \textbf{0.5194} & \textbf{0.4923} & \textbf{0.5462} & 0.5971 \\

\bottomrule

\end{tabular}}
\vspace{-10pt}
\end{table*}
\renewcommand{\arraystretch}{0.9}
\begin{table*}[t]
\centering
\caption{Performance variation according to the number of neighbors ($k$) in local attention. The best results are in \textbf{bold}.}
\label{tab:sup_k}

\setlength{\tabcolsep}{7pt}

\resizebox{\textwidth}{!}{%
\begin{tabular}{l ccc ccc ccc}

\toprule
\multirow{2}{*}{\textbf{$k$}} & \multicolumn{3}{c}{\textbf{ST-Net}} & \multicolumn{3}{c}{\textbf{Her2ST}} & \multicolumn{3}{c}{\textbf{SCC}} \\
\cmidrule(lr){2-4} \cmidrule(lr){5-7} \cmidrule(lr){8-10}

& \rule{1pt}{0ex} MSE $\downarrow$ & MAE $\downarrow$ & PCC $\uparrow$ \rule{1pt}{0ex} & \rule{1pt}{0ex} MSE $\downarrow$ & MAE $\downarrow$ & PCC $\uparrow$ \rule{1pt}{0ex} & \rule{1pt}{0ex} MSE $\downarrow$ & MAE $\downarrow$ & PCC $\uparrow$ \rule{1pt}{0ex} \\

\midrule
8   & 0.1138 & 0.2630 & 0.7176 & 0.6327 & 0.6440 & 0.4957 & 0.4920 & \textbf{0.5438} & 0.5975 \\
16  & 0.1119 & \textbf{0.2597} & \textbf{0.7237} & 0.6211 & 0.6366 & 0.5118 & \textbf{0.4887} & 0.5475 & \textbf{0.5987} \\
32  & \textbf{0.1111} & 0.2603 & 0.7224 & \textbf{0.6158} & 0.6342 & \textbf{0.5194} & 0.4923 & 0.5462 & 0.5971 \\
64  & 0.1140 & 0.2645 & 0.7177 & 0.6203 & 0.6379 & 0.5122 & 0.4902 & 0.5538 & 0.5980 \\
100 & 0.1122 & 0.2614 & 0.7207 & 0.6205 & \textbf{0.6321} & 0.5061 & 0.4904 & 0.5567 & 0.5979 \\
\bottomrule

\end{tabular}}
\vspace{-10pt}
\end{table*}

\renewcommand{\arraystretch}{0.9}
\begin{table*}[t!]
\centering
\caption{Impact of the temperature parameter ($\tau_{\text{neg}}$) on model performance. The best results are in \textbf{bold}.}
\label{tab:sup_tau}

\setlength{\tabcolsep}{7pt}

\resizebox{\textwidth}{!}{%
\begin{tabular}{l ccc ccc ccc}

\toprule
\multirow{2}{*}{\textbf{$\tau_{\text{neg}}$}} & \multicolumn{3}{c}{\textbf{ST-Net}} & \multicolumn{3}{c}{\textbf{Her2ST}} & \multicolumn{3}{c}{\textbf{SCC}} \\
\cmidrule(lr){2-4} \cmidrule(lr){5-7} \cmidrule(lr){8-10}

& \rule{1pt}{0ex} MSE $\downarrow$ & MAE $\downarrow$ & PCC $\uparrow$ \rule{1pt}{0ex} & \rule{1pt}{0ex} MSE $\downarrow$ & MAE $\downarrow$ & PCC $\uparrow$ \rule{1pt}{0ex} & \rule{1pt}{0ex} MSE $\downarrow$ & MAE $\downarrow$ & PCC $\uparrow$ \rule{1pt}{0ex} \\

\midrule
0.6 & \textbf{0.1111} & \textbf{0.2603} & \textbf{0.7224} & \textbf{0.6158} & \textbf{0.6342} & \textbf{0.5194} & \textbf{0.4923} & \textbf{0.5462} & \textbf{0.5971} \\
1.0 & 0.1128 & 0.2623 & 0.7191 & 0.6240 & 0.6374 & 0.5104 & 0.4926 & 0.5497 & 0.5944 \\
1.4 & 0.1129 & 0.2624 & 0.7212 & 0.6236 & 0.6371 & 0.5064 & 0.4966 & 0.5481 & 0.5952 \\
\bottomrule

\end{tabular}}
\vspace{-10pt}
\end{table*}

\renewcommand{\arraystretch}{0.9}
\begin{table*}[t]
\centering
\caption{Impact of the negative attention scaling factor ($\beta$) on model performance. The best results are in \textbf{bold}.}
\label{tab:sup_beta}

\setlength{\tabcolsep}{7pt}

\resizebox{\textwidth}{!}{%
\begin{tabular}{l ccc ccc ccc}

\toprule
\multirow{2}{*}{\textbf{$\beta$}} & \multicolumn{3}{c}{\textbf{ST-Net}} & \multicolumn{3}{c}{\textbf{Her2ST}} & \multicolumn{3}{c}{\textbf{SCC}} \\
\cmidrule(lr){2-4} \cmidrule(lr){5-7} \cmidrule(lr){8-10}

& \rule{1pt}{0ex} MSE $\downarrow$ & MAE $\downarrow$ & PCC $\uparrow$ \rule{1pt}{0ex} & \rule{1pt}{0ex} MSE $\downarrow$ & MAE $\downarrow$ & PCC $\uparrow$ \rule{1pt}{0ex} & \rule{1pt}{0ex} MSE $\downarrow$ & MAE $\downarrow$ & PCC $\uparrow$ \rule{1pt}{0ex} \\

\midrule
0.75 & \textbf{0.1108} & \textbf{0.2565} & \textbf{0.7253} & 0.6229 & 0.6358 & 0.5148 & \textbf{0.4907} & 0.5572 & \textbf{0.6002} \\
1.5  & 0.1111 & 0.2603 & 0.7224 & \textbf{0.6158} & \textbf{0.6342} & \textbf{0.5194} & 0.4923 & \textbf{0.5462} & 0.5971 \\
3.0  & 0.1155 & 0.2652 & 0.7088 & 0.6251 & 0.6397 & 0.5113 & 0.4969 & 0.5504 & 0.5952 \\
\bottomrule

\end{tabular}}
\vspace{-10pt}
\end{table*}

\renewcommand{\arraystretch}{0.9}
\begin{table*}[t]
\centering
\caption{Performance comparison between different feature extractors. The best results are in \textbf{bold}.}
\label{tab:sup_foundation}

\setlength{\tabcolsep}{7pt}

\resizebox{\textwidth}{!}{%
\begin{tabular}{l ccc ccc ccc}

\toprule
\multirow{2}{*}{Model} & \multicolumn{3}{c}{\textbf{ST-Net}} & \multicolumn{3}{c}{\textbf{Her2ST}} & \multicolumn{3}{c}{\textbf{SCC}} \\
\cmidrule(lr){2-4} \cmidrule(lr){5-7} \cmidrule(lr){8-10}

& \rule{1pt}{0ex} MSE $\downarrow$ & MAE $\downarrow$ & PCC $\uparrow$ \rule{1pt}{0ex} & \rule{1pt}{0ex} MSE $\downarrow$ & MAE $\downarrow$ & PCC $\uparrow$ \rule{1pt}{0ex} & \rule{1pt}{0ex} MSE $\downarrow$ & MAE $\downarrow$ & PCC $\uparrow$ \rule{1pt}{0ex} \\

\midrule
UNI2-h~\cite{uni} & \textbf{0.1111} & \textbf{0.2603} & \textbf{0.7224} & \textbf{0.6158} & \textbf{0.6342} & \textbf{0.5194} & 0.4923 & \textbf{0.5462} & 0.5971 \\
Prov-GigaPath~\cite{gigapath} & 0.1147 & 0.2645 & 0.7125 & 0.6279 & 0.6472 & 0.5121 & \textbf{0.4914} & 0.5468 & \textbf{0.5988} \\

\bottomrule
\end{tabular}}
\vspace{-10pt}
\end{table*}

\subsection{Pre-processing} \label{sec:details_preprocess}

\myparagraph{Smoothing.}
In this study, we applied the same SPCS~\cite{spcs} smoothing implementation and data preprocessing parameters as adopted in MERGE  for fair comparison. 
Specifically, we set the zero cutoff parameter for gene filtering to 0.7. 
According to the analysis in MERGE, it was reported that only an average of about 5.72 genes per sample are excluded when this criterion (0.7) is applied. 
Consequently, to prevent the inadvertent removal of potentially informative genes and to maintain consistent experimental conditions with MERGE, we omitted the separate gene filtering step. 
All other SPCS parameters were also kept exactly identical to the MERGE settings.

\myparagraph{Gene.}
All experimental evaluations were carried out based on the 250 target genes specified in TRIPLEX~\cite{triplex}. 
Detailed information regarding this specific gene set is provided in \Cref{tab:gene_list}.

\section{Extended Experimental Analysis}  \label{sec:analysis}

In this section, we provide additional experimental results and detailed analyses that complement the findings presented in the main paper. 
Specifically, to further demonstrate the robustness of our model's performance, we conducted experiments using different train/val/test splits of the same dataset used in the main paper.

\subsection{Ablation Study} \label{sec:analysis_ablation}

We present the complete ablation study results across all three benchmark datasets (ST-Net~\cite{stnet}, Her2ST~\cite{her2st}, and SCC~\cite{scc}) to validate our proposed components. 
\Cref{tab:sup_ablation_study} summarizes the performance changes when the negative-aware attention and off-grid sampling strategy are selectively applied. 
The baseline model, which utilizes neither component, generally exhibits the lowest performance across all metrics.
Introducing either component individually leads to performance gains compared to the baseline. 
Most importantly, our FEAST framework, which integrates both components, achieves the best or highly comparable results across all three datasets. 
For example, in the Her2ST dataset, the full model achieves the highest PCC of 0.5194, surpassing both the baseline (0.4920) and single-component variants.
These results confirm that the off-grid sampling strategy and negative-aware attention are complementary and essential for robust gene expression prediction in diverse tissue environments.

\begin{figure*}[t!]
    \centering
    \includegraphics[width=\textwidth]{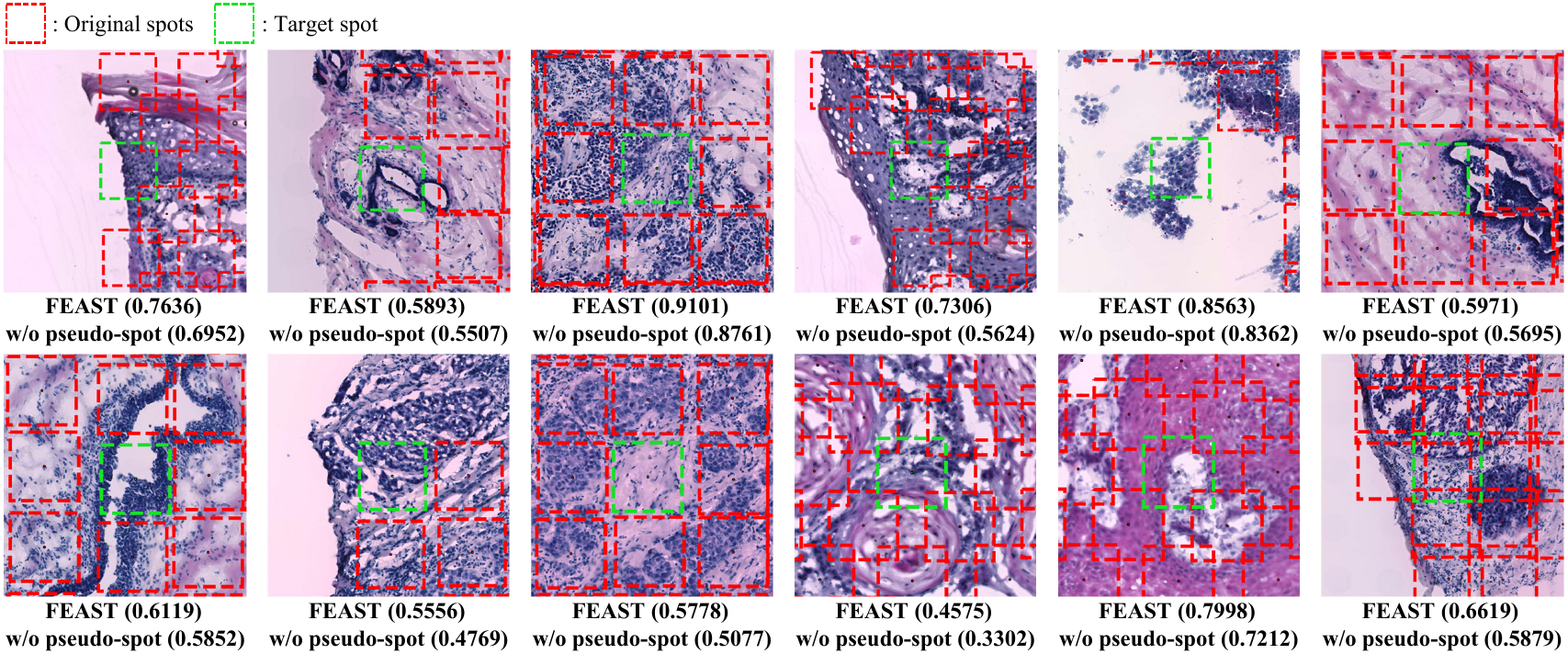}
    \vspace{-15pt}
    \caption{Additional illustration of the effect of pseudo-spots (off-grid sampling) on target spot prediction.}
    \label{fig:pseudo_supple}
    \vspace{-10pt}
\end{figure*}

\subsection{Detailed Analysis of Design Elements} \label{sec:analysis_elements}

In addition to the ablation study, we conducted a series of experiments to analyze the impact of various hyperparameters and design choices.

\myparagraph{Effect of Number of Neighbors ($k$).}  In the main paper, we reported the impact of the number of neighbors $k$ in local $k$-NN Attention only for the Her2ST dataset. 
Here, we expand this analysis to all three datasets, with results summarized in \Cref{tab:sup_k}. 
The results illustrate that performance generally improves as $k$ increases, with the most robust results observed in the range of $k=16$ to $32$. 
Specifically, while $k=16$ achieved the best performance in several metrics for ST-Net and SCC, $k=32$ yielded the optimal results for Her2ST and remained highly competitive across the other datasets. 
Notably, further increasing the number of neighbors to $k=64$ or $100$ did not lead to consistent performance gains and, in several instances, resulted in marginal degradation. 
This suggests that an excessively large window may introduce noise from biologically distant spots. 
Considering the stable performance across all benchmarks and the computational efficiency, we selected $k=32$ as the default value for FEAST.

\myparagraph{Analysis of Temperature Scaling ($\tau_{\text{neg}}$).} We analyzed the variations in model performance with respect to the temperature parameter $\tau_{\text{neg}}$ in negative-aware attention. 
\Cref{tab:sup_tau} presents the results for $\tau_{\text{neg}} \in \{0.6, 1.0, 1.4\}$. 
The experimental results show that $\tau_{neg}=0.6$ consistently yields the best performance across all three datasets (ST-Net, Her2ST, and SCC) and all evaluation metrics. 
Specifically, increasing the temperature to 1.0 or 1.4 led to a general decline in both error metrics and correlation scores. 
This suggests that a lower temperature effectively sharpens the attention distribution and forces the model to focus only on the strongest negative relationships. 
Accordingly, we selected $\tau_{neg}=0.6$ as the default value for FEAST, given its robust and superior performance across diverse tissue environments.

\myparagraph{Impact of Negative Attention Scaling ($\beta$).} We evaluated the impact of the hyperparameter $\beta$, which scales the contribution of the negative attention weights in our final attention computation. 
\Cref{tab:sup_beta} presents the performance variations across different $\beta$ values. 
The results indicate that the optimal $\beta$ value varies across datasets; for instance, while $\beta=0.75$ achieved the highest PCC scores for ST-Net (0.7253) and SCC (0.6002), $\beta=1.5$ yielded the most robust performance for the Her2ST dataset. 
Notably, increasing $\beta$ to 3.0 led to a consistent decline in performance across all benchmarks, suggesting that an excessive negative contribution may overshadow essential positive spatial relationships. 
Although we fixed $\beta=1.5$ as the default for consistency in the main paper, these findings indicate that dataset-specific fine-tuning could yield further performance improvements.

\myparagraph{Comparison of Feature Extractors.}
We compared UNI2-h~\cite{uni} and Prov-GigaPath~\cite{gigapath} as feature extractors for FEAST. 
As shown in \Cref{tab:sup_foundation}, UNI2-h outperformed Prov-GigaPath in the majority of evaluation metrics, achieving the best results in 7 out of 9 cases across all datasets. 
While Prov-GigaPath showed marginal advantages in specific SCC metrics, UNI2-h demonstrated superior overall performance, particularly on the ST-Net and Her2ST. 
Consequently, we adopted UNI2-h as the default backbone for our framework due to its robust performance across most benchmarks.

\section{Additional Visualization} \label{sec:visualization}

\subsection{Efficacy of Off-grid Sampling} \label{sec:visualization_pseudo}

\Cref{fig:pseudo_supple} illustrates the comparative performance (PCC) of gene expression prediction with and without the application of pseudo-spots. 
The results demonstrate that incorporating pseudo-spots consistently leads to higher PCC values. 
Notably, this performance gain is most pronounced in challenging scenarios characterized by insufficient morphological context, such as regions with spatially sparse neighbors, tissue boundaries, or spots exhibiting structural patterns distinct from their surroundings.

\subsection{Visualization of Negative-aware Attention} \label{sec:visualization_attn}

\begin{figure*}[t!]
    \centering
    \includegraphics[width=\textwidth]{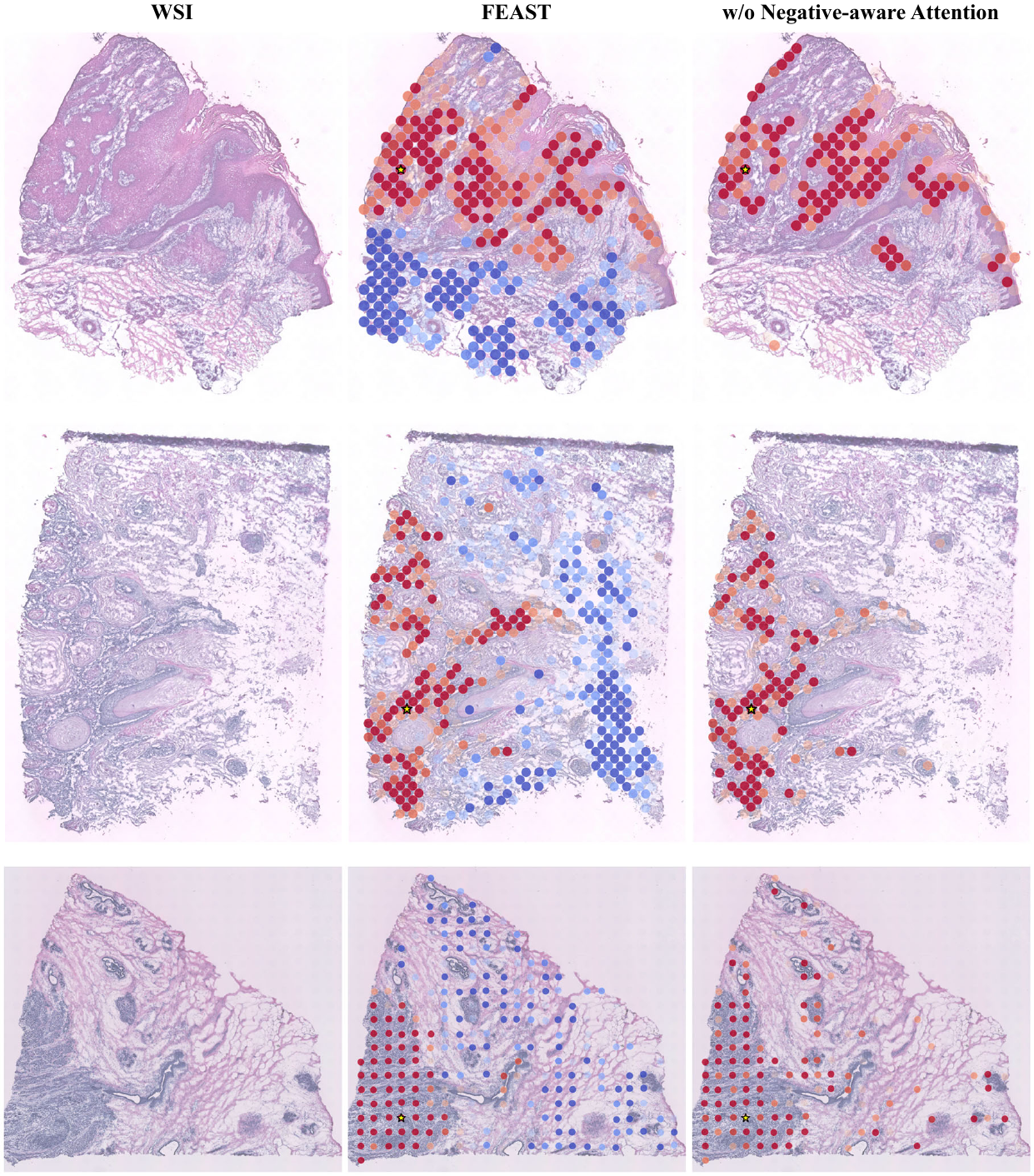}
    \vspace{-15pt}
    \caption{Comparison of attention maps with and without negative-aware attention.}
    \label{fig:attention_1}
    \vspace{-10pt}
\end{figure*}

\begin{figure*}[t!]
    \centering
    \includegraphics[width=\textwidth]{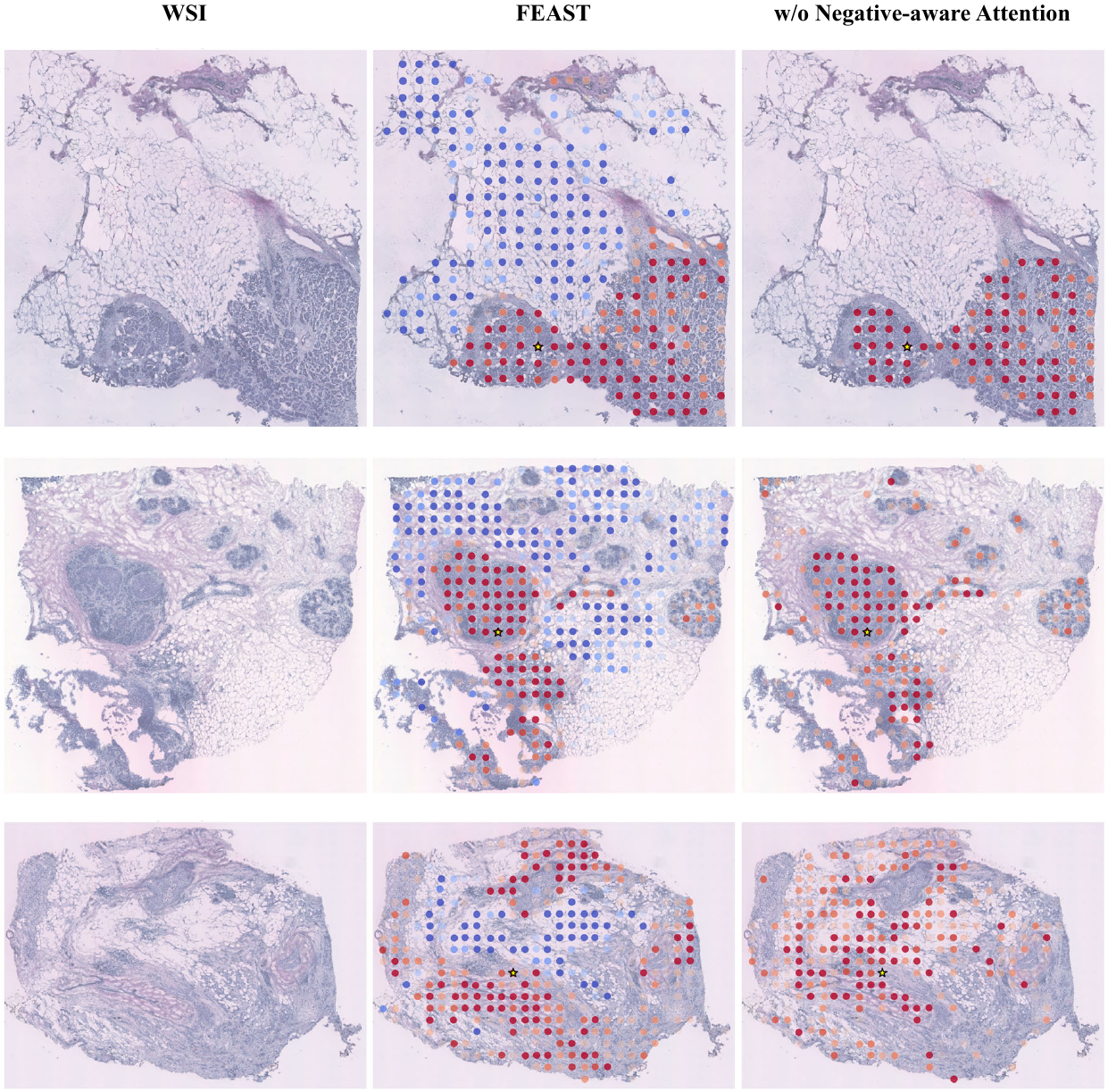}
    \vspace{-15pt}
    \caption{Comparison of attention maps with and without negative-aware attention.}
    \label{fig:attention_2}
    \vspace{-10pt}
\end{figure*}

\Cref{fig:attention_1} and \Cref{fig:attention_2} visualize the attention maps for the target spot with and without the use of negative-aware attention. With the introduction of this mechanism, the model assigns positive weights to biologically similar regions while simultaneously assigning negative weights to tissues with distinct gene expression patterns. This demonstrates that the model does not indiscriminately assign positive weights based on spatial proximity, but rather accurately discriminates between positive and negative relationships.

\subsection{PCC Comparison} \label{sec:visualization_PCC}

\begin{figure*}[t!]
    \centering
    \includegraphics[width=\textwidth]{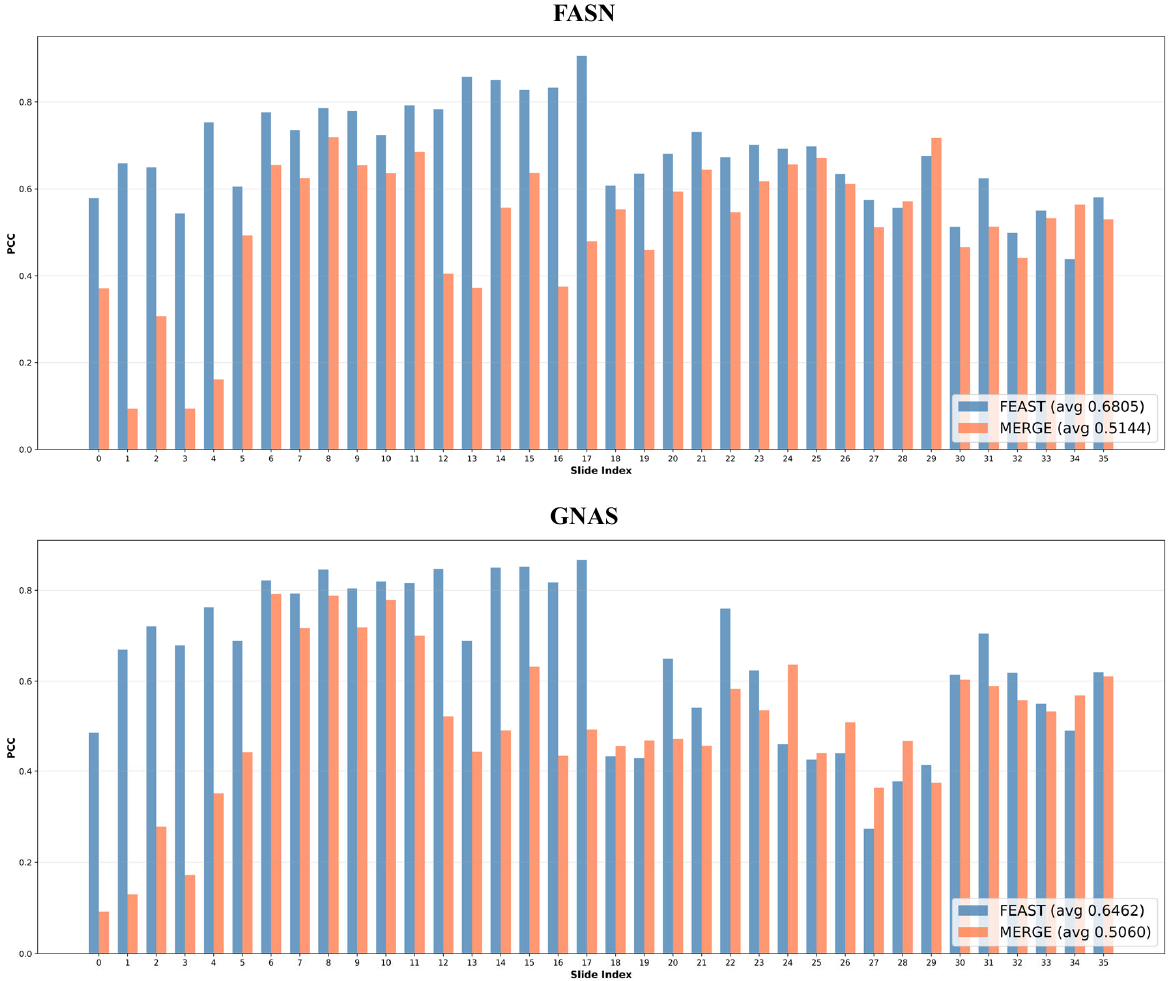}
    \vspace{-15pt}
    \caption{Bar plots of PCC scores for predicted FASN and GNAS gene expressions for each sample in the ST-Net dataset. The average PCC across the entire dataset is indicated in parentheses within the legend.}
    \label{fig:pcc_slide}
    \vspace{-10pt}
\end{figure*}

We conducted a comparative analysis on the Her2ST dataset to evaluate the PCC performance for two key cancer-associated genes, FASN and GNAS~\cite{fasn, gnas}. 
As illustrated in \Cref{fig:pcc_slide}, FEAST demonstrated a robust correlation capability, consistently outperforming MERGE across the majority of slide samples. 
While the legend indicates the superior average PCC of our method, a more granular look reveals that FEAST maintains high correlation scores even in instances where MERGE yields negative values. 
Quantitatively, FEAST surpassed the baseline in 33 and 28 out of the 36 total samples for FASN and GNAS, respectively.

\begin{figure*}[t!]
    \centering
    \includegraphics[width=\textwidth]{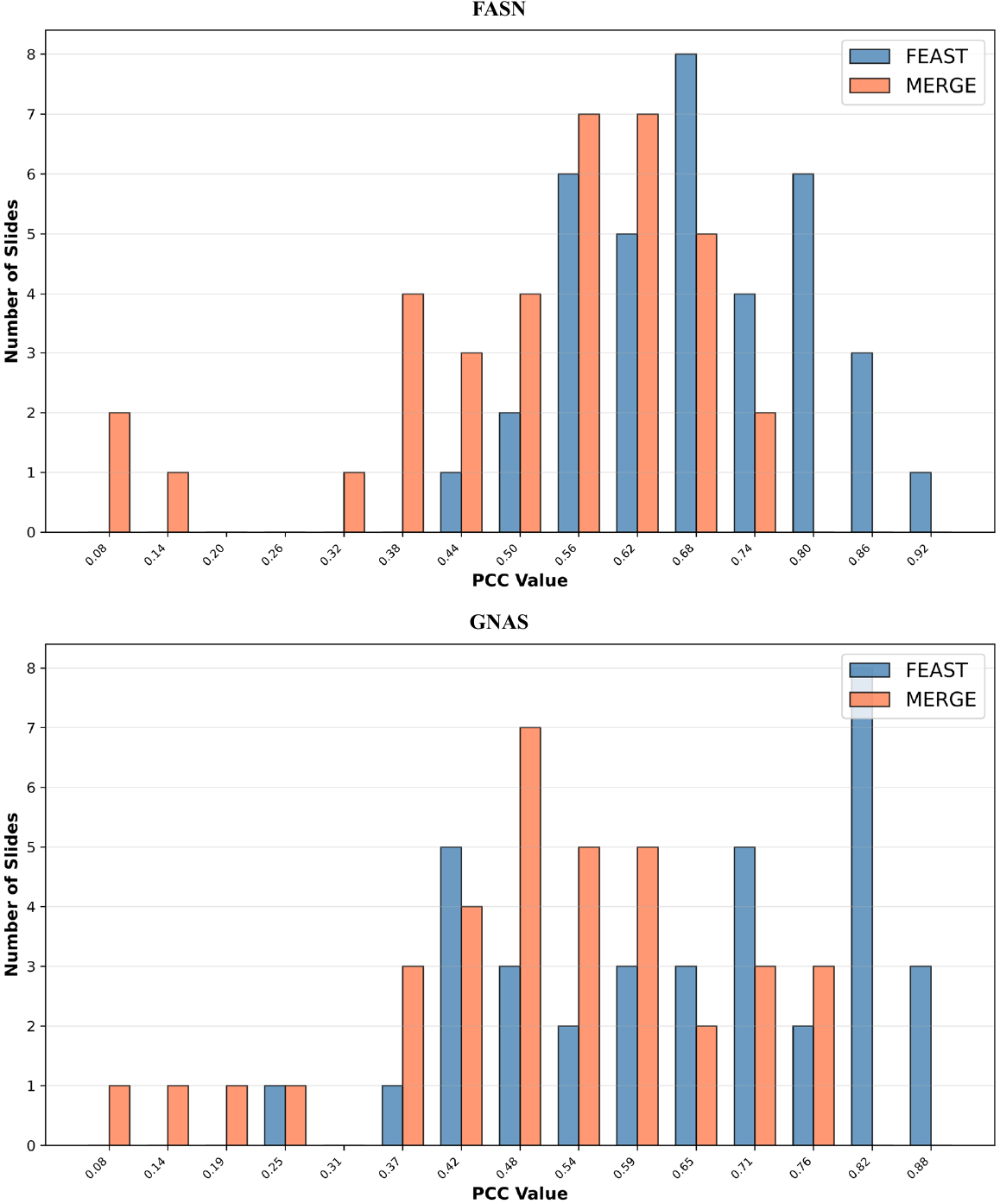}
    \vspace{-15pt}
    \caption{Histograms of PCC scores for FASN and GNAS predictions by the two methods on the ST-Net dataset. The distributions show that FEAST achieves higher PCC values compared to MERGE. Notably, it is evident that FEAST rarely yields negative PCC values.}
    \label{fig:pcc_histogram}
    \vspace{-10pt}
\end{figure*}

This performance gap is visually summarized in \Cref{fig:pcc_histogram}. 
The PCC distribution for FEAST is concentrated in the positive high-value region, standing in contrast to MERGE. 
This rightward shift in the distribution highlights FEAST's ability to consistently generate accurate gene expression predictions while minimizing instances of negative PCC values.

\subsection{Gene Expression Comparison} \label{sec:visualization_gene}

\begin{figure*}[t!]
    \centering
    \includegraphics[width=\textwidth]{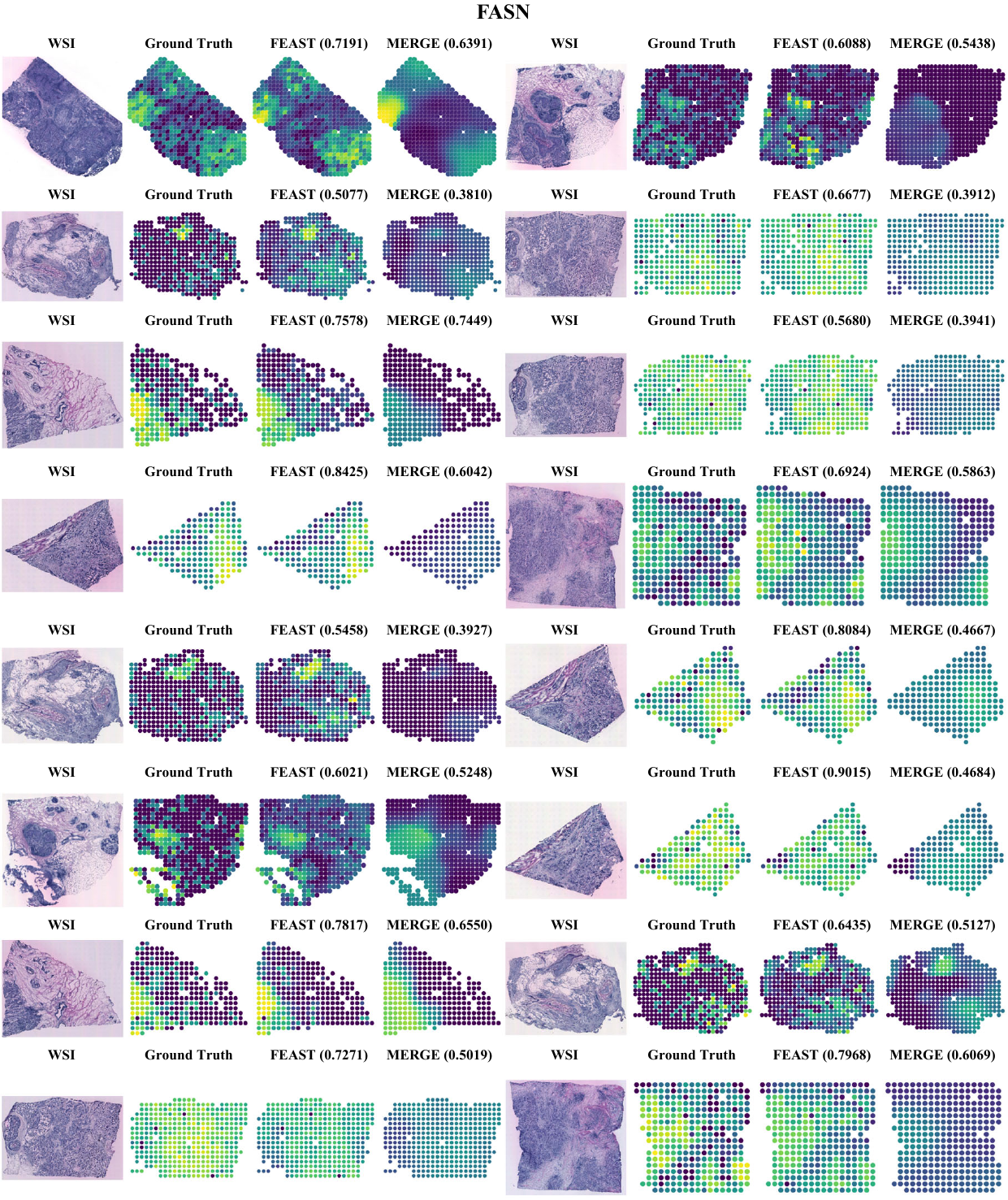}
    \vspace{-15pt}
    \caption{Qualitative comparison of predicted FASN gene expression heatmaps.}
    \label{fig:FASN}
    \vspace{-10pt}
\end{figure*}

\begin{figure*}[t!]
    \centering
    \includegraphics[width=\textwidth]{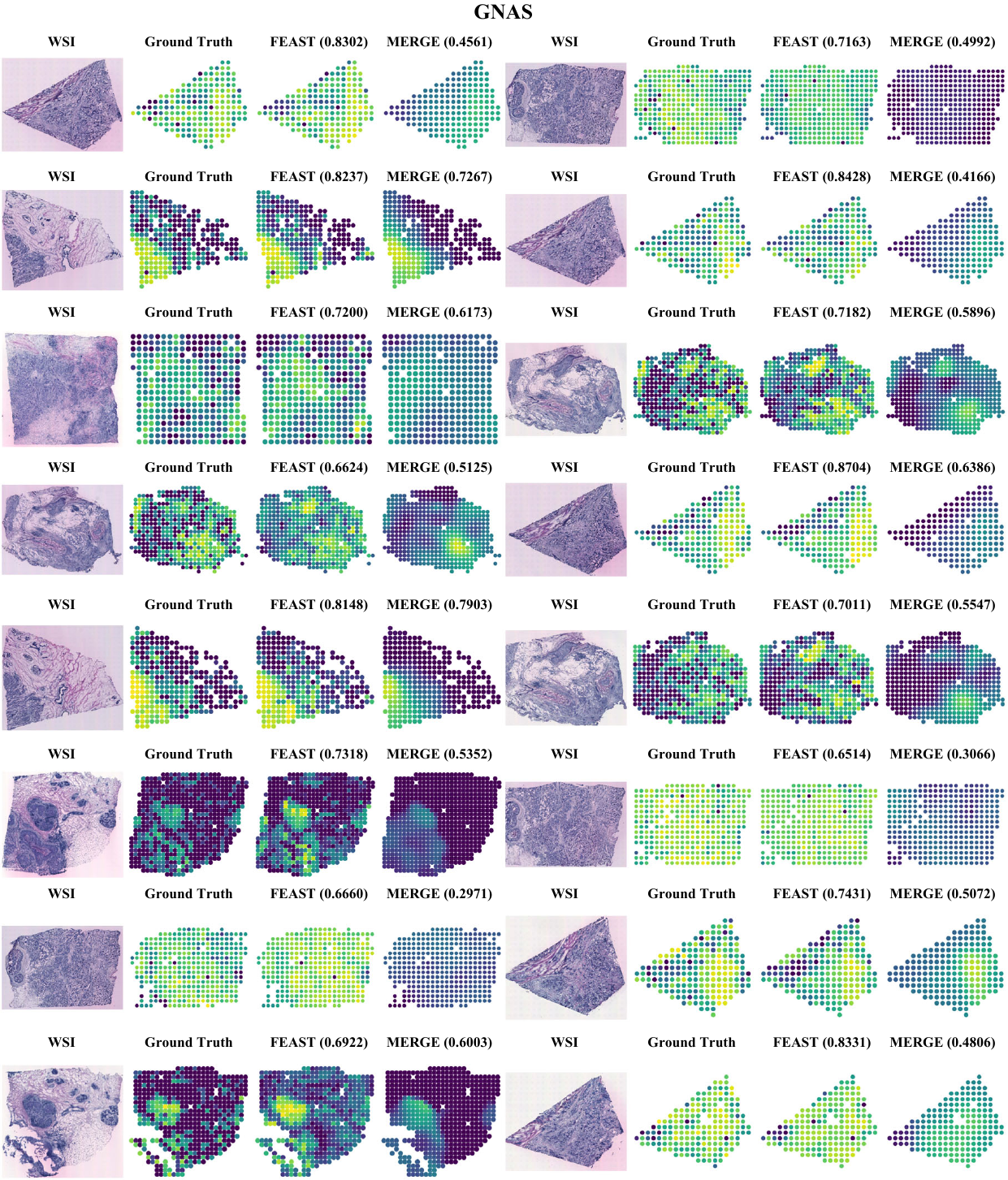}
    \vspace{-15pt}
    \caption{Qualitative comparison of predicted GNAS gene expression heatmaps.}
    \label{fig:GNAS}
    \vspace{-10pt}
\end{figure*}

\Cref{fig:FASN} and \Cref{fig:GNAS} present the predicted gene expression heatmaps for FASN and GNAS, respectively, comparing FEAST and MERGE against the ground truth. 
Visual inspection reveals that FEAST generates predictions that are significantly closer to the ground truth than those of MERGE. 
Notably, FEAST accurately captures individual spots with expression characteristics distinct from their surroundings, demonstrating the superior spatial resolution of our model.

\end{document}